\documentclass{article}

\usepackage{microtype}
\usepackage{graphicx}
\usepackage{subcaption}
\usepackage{booktabs} % for professional tables

\usepackage{hyperref}

\usepackage[preprint]{icml2026}

\usepackage{amsmath}
\usepackage{amssymb}
\usepackage{mathtools}
\usepackage{amsthm}

\usepackage[capitalize,noabbrev]{cleveref}
\theoremstyle{plain}

\theoremstyle{definition}

\theoremstyle{remark}

\usepackage[textsize=tiny]{todonotes}

\usepackage{caption}
\usepackage{bbm}
\usepackage{wrapfig}
\icmltitlerunning{LAVA: Explainability for Unsupervised Latent Embeddings}

\begin{document}

\twocolumn[
  \icmltitle{LAVA: Explainability for Unsupervised Latent Embeddings}



  \icmlsetsymbol{equal}{*}

  \begin{icmlauthorlist}
    \icmlauthor{Ivan Stresec}{delft}
    \icmlauthor{Joana P. Gonçalves}{delft}
  \end{icmlauthorlist}

  \icmlaffiliation{delft}{Delft University of Technology, The Netherlands}

  \icmlcorrespondingauthor{Ivan Stresec}{i.stresec@tudelft.nl}
  \icmlcorrespondingauthor{Joana P. Gonçalves}{joana.goncalves@tudelft.nl}

  \icmlkeywords{Machine Learning, ICML}

  \vskip 0.3in
]



\printAffiliationsAndNotice{}  % no special notice (required even if empty)

\begin{abstract}
  Unsupervised black-box models are drivers of scientific discovery, yet are difficult to interpret, as their output is often a multidimensional embedding rather than a well-defined target. While explainability for supervised learning uncovers how input features contribute to predictions, its unsupervised counterpart should relate input features to the structure of the learned embeddings. However, adaptations of supervised model explainability for unsupervised learning provide either single-sample or dataset-summary explanations, remaining too fine-grained or reductive to be meaningful, and cannot explain embeddings without mapping functions. To bridge this gap, we propose LAVA, a post-hoc model-agnostic method to explain local embedding organization through feature covariation in the original input data. LAVA explanations comprise modules, capturing local subpatterns of input feature correlation that reoccur globally across the embeddings. LAVA delivers stable explanations at a desired level of granularity, revealing domain-relevant patterns such as visual parts of images or disease signals in cellular processes, otherwise missed by existing methods. 
\end{abstract}
%/robust/consistent

\section{Introduction}
Pragmatically, an explanation can be defined as an answer to a ``why'' question~\cite{fraassen_pragmatic_1988}. Traditional models, like Newton's law of gravitation, constitute explanations as they allow reasoning about observable phenomena through their rigorous mathematical description. Similarly, stochastic phenomena that lend themselves well to statistical modeling, such as linear regression, are explainable through probabilistic reasoning.
For more complex problems such as facial recognition or protein structure prediction, expert-led modeling is superseded by machine learning (ML), where flexible model architectures and automated end-to-end optimization algorithms take center stage. ML has seen considerable progress and success in leveraging high-dimensional data for scientific discovery across fields like chemistry, mathematics, and biology~\cite{jumper_highly_2021, davies_advancing_2021, movva_deciphering_2019, ching_opportunities_2018}. However, even when ML models make reliable predictions or reflect observed phenomena, reasoning about how they do so is notoriously difficult as most are ``black boxes'': the complexity of the data and operations required to produce an output often exceeds what is practically comprehensible~\cite{lipton_mythos_2017, humphreys_philosophical_2009, beisbart_philosophy_2022}. %: even with full decomposibility
Limited insight into model behavior is a major barrier to adoption of ML for decision making, especially in critical areas such as healthcare~\cite{rudin_stop_2019, murdoch_definitions_2019, ching_opportunities_2018, wang_scientific_2023}.
To improve understanding of ML, the community uses either intrinsically interpretable but restrictive models, or explainability methods that allow for post-hoc interpretation of black-box models%or \textit{post-hoc} methods enabling explainability of black-box models
~\cite{rudin_stop_2019, higgins_towards_2018, scholkopf_causality_2019, murdoch_definitions_2019, molnar2022}. 

The popularity and complexity of black-box ML poses challenges to existing explainability methods, creating a need for new solutions. At its core, \textit{post-hoc} explainability aims to enable reasoning about the output of a model in terms of its treatment of the input data.
This allows cross-referencing of model behavior with domain knowledge for verification, assessment of model reliability, and mitigation of unwanted biases.
Existing explainability methods cover a wide range of approaches, including feature or sample importance, anchors, and counterfactual explanations, among others~\cite{molnar2022}.
However, most are developed to explain supervised ML models, whose output tends to be a well-defined target label or quantity. This ignores unsupervised black-box ML, designed to reveal latent structure with limited prior knowledge, which can contribute meaningfully to scientific discovery. Indeed, latent embeddings based on autoencoder and transformer architectures have shown effectiveness in various tasks~\cite{brown_language_2020, ramesh_zero-shot_2021, makrodimitris_-depth_2024}, some serving as a basis of foundation models~\cite{bommasani_opportunities_2022}. % mention examples of tasks (vision, natural language processing, proteins, etc.)?
Likewise, manifold learning is prominently used in bioinformatics for data exploration, trajectory inference, cell type prediction, spatial domain identification, and multiomics integration~\cite{kobak_art_2019, brbic_annotation_2022, saelens_comparison_2019, wolf_paga_2019, haviv_covariance_2024}. 

Crucially, most unsupervised black-box models output multi-dimensional latent embeddings. Relating latent dimensions to the input data is less straightforward, as the axes of the learned latent space do not correspond to known quantities and are therefore difficult to interpret~\cite{crabbe_label-free_2022}. Disentangled representations have shown mixed results, %, and might altogether be impossible without supervision based on prior knowledge~\cite{crabbe_label-free_2022, locatello_sober_2020}, 
suggesting that latent dimensions cannot generally be explained independently from one another~\cite{crabbe_label-free_2022, locatello_sober_2020}. As such, a promising direction for explainability of unsupervised black-box models is to focus on the relation between the relative spatial organization of the embeddings and the underlying data. Moreover, many manifold learning methods only produce latent representations of the samples, without a parametrized mapping function to embed new data~\cite{murphy_probabilistic_2022}. This means that all information is contained within the embeddings, making the link between latent organization and underlying data crucial for reasoning about the model.

Literature on explainability of unsupervised black-box latent embeddings is limited. Most notably, \citet{crabbe_label-free_2022} propose a framework adapting feature and sample importance explainability for supervised models to the unsupervised setting. This contribution importantly enables explanations for unsupervised ML using a wide range of methods. However, these yield either single-sample or dataset-summary explanations, which ignore the organization of latent embeddings and are typically too fine-grained or too global to meaningfully understand the underlying model. This is especially relevant as many black-box models are non-linear and seek to capture local structure in the original high-dimensional data. In fact, the popular t-SNE and UMAP manifold learning algorithms ~\cite{meila_manifold_2024, maaten_visualizing_2008, mcinnes_umap_2020} explicitly aim to preserve local sample neighborhoods of the input in a lower-dimensional latent space. We therefore argue that effective explainability of latent embeddings requires some strategy of relating samples to one another based on their organization in the latent space. While manifold learning methods have been extensively studied in terms of neighborhood retention~\cite{han_enhance_2022} and hyperparameter effects%further studied to recommend ways of setting hyperparameters
~\cite{becht_dimensionality_2019, kobak_art_2019, wang_understanding_2021, xia_statistical_2024}, no research has focused on explaining the organization of the embeddings as a function of the input data. %While these studies offer insight into these methods, they do not provide information on how the embeddings relate to the original data. 
More manual approaches relate spatial organization to original features~\cite{liu_latent_2019}, but relying purely on prior knowledge, metadata, or target features of interest, and are infeasible for high-dimensional data.

Here, we propose Locality-Aware Variable Associations (LAVA), a post-hoc model-agnostic method to explain learned latent embeddings by relating input features to latent spatial organization. 
%To explain the embeddings, LAVA extracts modules that capture reoccurring subpatterns of input feature correlations throughout localities representing sample neighborhoods in the latent space. LAVA explanations therefore comprise both the modules and their presences in localities, describing the relation between local latent structure and input feature covariation in the underlying data.
LAVA generates explanations as modules capturing local subpatterns of input feature correlation reoccurring globally in representative neighborhoods throughout the latent space. LAVA modules therefore describe latent structure at a desired level of granularity, in terms of input feature covariation in the underlying data. 
%LAVA explanations take the form of modules, globally reoccurring subpatterns of input feature correlation found throughout representative embedding neighborhoods, called localities. 
%LAVA explanations therefore describe embeddings in terms of global correlation patterns of the input features, observed locally in the latent organization/structure.
%LAVA explains the input feature covariation across the latent space using modules, reoccurring subpatterns of input feature correlations found throughout the latent space. LAVA extracts modules based on localities, neighborhoods that serve as a reduced representation of the latent embeddings. The modules, together with presence values that quantify in which localities the modules are present in, serve as explanations and
% elucidate
%describe the relationship between local latent structure and the features of the underlying data. 
We use LAVA to explain embeddings of the MNIST and single-cell kidney datasets, showcasing its ability to identify visually and biologically relevant modules of pixel and gene expression correlation, respectively. %Through several experiments, we also demonstrate the stability of our method, the properties of the extracted modules, and how our method differs from existing approaches.
We further characterize LAVA explanations and assess their stability. Finally, we also compare LAVA to alternative explainability and module extraction methods. 

% Old version of last paragraph
% To explain local structure, LAVA creates explanations based on localities: neighborhoods of sample embeddings across the latent space. Each latent locality is linked to the original data through a representation comprising all pairwise correlations between the original features, considering only the original samples belonging to that locality's neighborhood. To reveal the consistency of local relationships between localities throughout the latent space, LAVA identifies modules of reoccurring subpatterns of feature correlations across localities. Experiments with LAVA reveal that neighborhoods of embeddings across a latent space can show distinct correlation patterns but also share subpatterns of correlation, and demonstrate that LAVA modules capture similar subpatterns between localities. We use LAVA to explain UMAP embeddings of both MNIST and single-cell kidney datasets, and showcase its ability to identify visually and biologically relevant modules of pixels and gene expression correlation, respectively.

\begin{figure*}[ht]
    \centering
    \includegraphics[width=\linewidth]{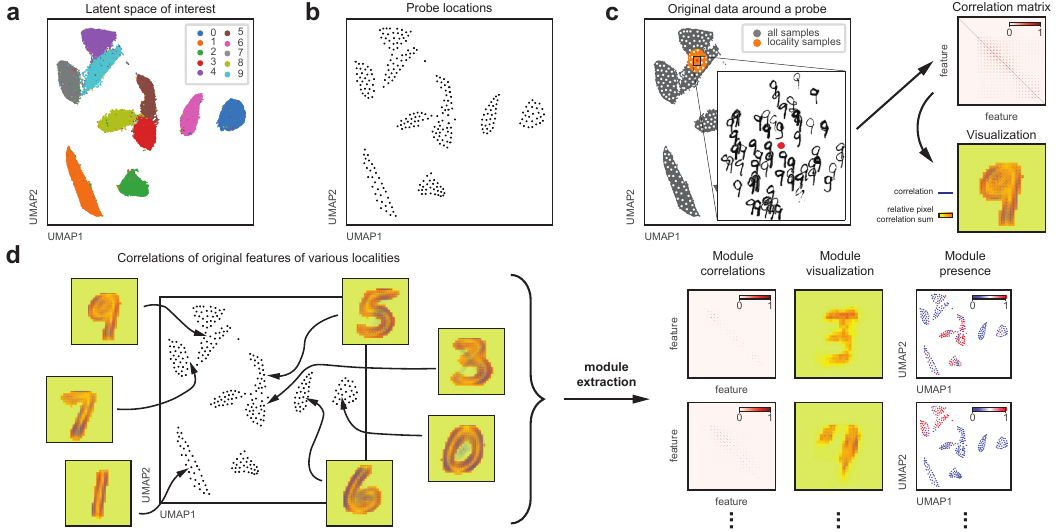}
    \caption{\textbf{LAVA method overview.} \textbf{a}, LAVA explains embeddings of a dataset (example UMAP embeddings of MNIST). \textbf{b}, Based on neighborhood size $n$, LAVA represents embeddings as a set of localities: overlapping neighborhoods centered around probes. \textbf{c}, For each locality, LAVA calculates feature-to-feature correlations using the original samples of that locality (top right, locality correlations; bottom right, heatmap showing the sum of correlations per pixel, %highlighting features or pixels contributing to a large number of correlations, 
    with added lines connecting correlated pixels). %showing stronger correlations. 
    \textbf{d},~LAVA extracts modules, shared subpatterns of correlation extracted from localities that serve as explanations, describing feature covariation across the embeddings.}
    \label{fig:method_overview}
\end{figure*}

\section{Methodology}
\label{sec:methods}
LAVA creates explanations relating local organization of latent embeddings to feature covariation of the underlying data in three steps: %The LAVA method consists of three interdependent steps: 
(1) locality definition, (2) locality representation, and (3) module extraction (Figure~\ref{fig:method_overview}). 
Step 1 defines the scale and reference points (probes) of the neighborhoods, or localities, to use for capturing local organization. 
%Step 1 defines at what scale the local organization will be investigated and decides which reference points (probes) should be used to do so. 
Step 2 creates a representation of each locality, %ies around every probe, 
linked to the original features via their correlation. Step 3 produces the actual explanations, consisting of modules of common correlation subpatterns across the localities.

\subsection{Locality Definition}
To reflect the local organization of latent embeddings, LAVA uses a smaller number of localities, centered around probe points (not samples) spread across the latent space (Figure~\ref{fig:method_overview}b). A locality centered at probe $p$ is an $n$-sized Euclidean neighborhood containing the $n$ samples nearest to $p$ in the latent space. The number of localities $\ell \leq E$ is given by $\ell = E \times o/n$, based on the number of embeddings $E$, neighborhood size $n$, and factor $o$ controlling the extent of overlap between localities (Appendix~\ref{app:locality_definition}).

\textbf{Locality placement.}
The goal of locality placement is to determine the locations of the $\ell$ probes around which the $\ell$ localities will be formed. To ensure that the localities are representative of the latent embedding structure, LAVA optimizes locality in-degree centrality towards neighborhood in-degree centrality (Figures~\ref{fig:mnist_probes}, \ref{fig:mnist_probes_ae}, and \ref{fig:kpmp_probes}). 
%LAVA determines the placement of localities by optimizing locality in-degree centrality towards neighborhood in-degree centrality. 
In other words, the proportion of probe-centered localities including a given sample  is made to approximate the proportion of all possible sample-centered $n$-neighborhoods including that same sample. 
%In other words, by making the number of probe-based localities including a sample as a neighbor approximate the number of $n$-neighborhoods including a sample as a neighbor when considering all sample-based $n$-sized neighborhoods, in relative terms (Figures~\ref{fig:mnist_probes}, \ref{fig:mnist_probes_ae}, and \ref{fig:kpmp_probes}). 
Practically, LAVA finds the locations of the $\ell$ probes as cluster centers determined using a custom weighted $k$-means formulation, where $k=\ell$. %(Appendix~\ref{app:locality_definition}).
%Practically, locality placement is determined based on weighted $k$-means clustering, using the DIRECT optimization algorithm~\cite{macqueen_methods_1967, jones_lipschitzian_1993}. The goal is to find the locations of the $\ell$ probes as cluster centers in $k$-means ($k=\ell$), around which the $\ell$ localities will be formed. 
The clustering uses the following sample weighting scheme:
\begin{equation}
    \centering
        w_i =  \left(\frac{in\_neighbor_i}{E}\right)^{\alpha} \times \left(\frac{1}{\bar{d}_i^n} \right)^{\beta},
    \label{eq:weighing_scheme}
\end{equation}  
where $w_i$ is the weight of sample $i$, $in\_neighbor_i$ is the in-degree centrality of sample $i$ using sample-based $n$-neighborhoods, $\bar{d}_i^n$ is the average distance from sample $i$ to its $n$ nearest neighbors, and $\alpha$ and $\beta$ are the parameters optimized using the DIRECT algorithm~\cite{macqueen_methods_1967, jones_lipschitzian_1993}. %Appendix~\ref{app:locality_definition}). 
We define the cost function for DIRECT as the Manhattan distance between the neighborhood and locality in-degree centralities of each sample, as follows:
    \begin{equation}
        \centering
            \mathcal{L}(\alpha, \beta) = \sum_{i=1}^{E}\left|\frac{in\_neighbor_i}{E} - \frac{in\_locality_{i,\alpha, \beta}}{\ell}\right|,
        \label{eq:direct_loss}
    \end{equation}
where $in\_locality_{i, \alpha, \beta}$ is the number of localities around the resulting $\ell$ cluster centers that include $i$ as a neighbor. Details and pseudocode in Appendix \ref{app:locality_definition}.

\subsection{Locality Representation}
After defining localities, LAVA links these local organizations of embeddings to the original data. This is achieved by representing each locality through pairwise associations of the original features, calculated using only the samples within that locality. In this version of LAVA, association is defined as the absolute Spearman correlation, capturing the strength of monotone association between features as values in $[0, 1]$ (Figure~\ref{fig:method_overview}c). Therefore, each locality is represented by a vector of $D \choose 2$ pairwise correlations, where $D$ is the number of features. Unlike Pearson correlations, Spearman correlations are not scale dependent and thus more flexible in the type of association they capture, while remaining computationally inexpensive and interpretable downstream. Individual correlations are filtered out (set to $0$) if one of the original features has the same value in $>75\%$ of the neighborhood (Appendix~\ref{app:calculating_correlations}).

\begin{figure*}[ht]
    \centering
    \includegraphics[width=\linewidth]{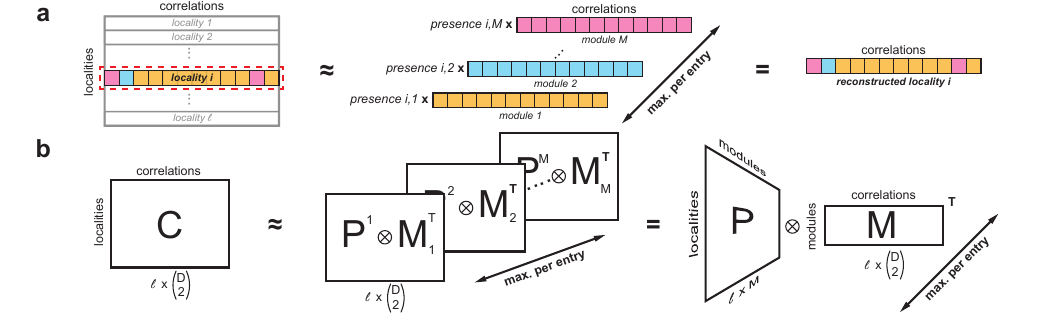}
    \caption{\textbf{LAVA association matrix factorization (AMF).} %Module-based factorization of localities.} 
    \textbf{a}, Locality $i$ is reconstructed by multiplying each module by its presence for $i$, then taking the maximum per entry across the $M$ extracted modules. 
    \textbf{b}, Localities are represented as matrix $\boldsymbol{C}$: each row is a locality, each column a pair of features, and each entry the pairwise feature correlation in that locality. Module extraction approximates $\boldsymbol{C}$ with a maximum per entry based on modules $\boldsymbol{M}$ and their presences $\boldsymbol{P}$: it calculates an outer product between each column of $\boldsymbol{P}$ (superscript) and each row of $\boldsymbol{M}$, yielding a tensor of $M$ matrix slices of the same shape as $\boldsymbol{C}$ (Equation \ref{eq:module_outer_product}). Each slice represents module-contributed correlations across the localities. The $M$ dimension reduced by taking the per element maximum yields the reconstructed locality dataset.} 
    \label{fig:module_extraction}
\end{figure*}

\subsection{Module Extraction}
\label{sec:module_extraction}

Finally, LAVA extracts modules of reoccurring feature correlation subpatterns across localities, as explanations of the latent embeddings (Figure~\ref{fig:module_extraction}). Modules adhere to two properties: (1) modules capture reoccurring rather than average patterns across localities, as modules based on averages of localities would not be accurate representations of correlations observed at each locality individually; (2) modules can be co-present in the same locality, allowing decomposition of the correlation patterns. %into smaller subpatterns.

Modules are identified based on the matrix of locality representations $\boldsymbol{C}$, with rows denoting the $\ell$ localities and columns representing the ${D \choose 2}$ pairwise correlations. To identify modules, we propose the Association Matrix Factorization (AMF) method, which factorizes a matrix like $\boldsymbol{C}$ into two module matrix $\boldsymbol{M}$ and presence matrix $\boldsymbol{P}$. Module matrix $\boldsymbol{M}$ is of size $M \times {D \choose 2}$, where each row represents one of $M$ modules, columns denote correlations, and each row vector expresses a subpattern of correlations of a module. Presence matrix $\boldsymbol{P}$ is of size $\ell \times M$, with each entry quantifying the presence of one of the $M$ modules in one of the $\ell$ localities. Presence and module matrix entries are restricted to real values in $[0, 1]$, and reconstruct locality correlations using the formulation below. 

\textbf{AMF formulation.} The formulation of AMF enables modules to overlap across localities. Crucially, as correlations or measures of association more broadly cannot be considered additive, we avoid summing correlations and instead use a per element maximum function. This allows modules to contribute overlapping subpatterns of correlations without direct interaction in the reconstruction. To that effect, we reconstruct each correlation $j$ of each locality $i$ by taking the maximum of the modules' corresponding correlation values multiplied by the presences of those modules at locality $i$:
\begin{equation}
    \centering
        \boldsymbol{\hat{C}}_{ij} = \max\left(\boldsymbol{P}_{i} \odot (\boldsymbol{M}^T)_{j}\right),
    \label{eq:correlation_module_formulation}
\end{equation}
where $\odot$ is the Hadamard product and single letter subscripts indicate matrix rows. The reconstruction of a locality $i$ can then be written as a per element maximum across products of the modules and their presences in locality $i$ %locality's presence values and the modules %i$ 
(Figure~\ref{fig:module_extraction}a):
\begin{equation}
    \centering
        \boldsymbol{\hat{C}}_{i} = \max\limits_{1 \leq j \leq {D \choose 2}}\left(\boldsymbol{P}_{i1} \times \boldsymbol{M}_{1}^T, ..., \boldsymbol{P}_{iM} \times \boldsymbol{M}_{M}^T\right),
    % \boldsymbol{P}_{i2} \times \boldsymbol{M}_{2}^T, 
    \label{eq:locality_module_formulation}
\end{equation}
where $j$ is the index of a single correlation. Similarly, across all $\ell$ localities, we can take the per element maximum of $M$ matrices (Figure \ref{fig:module_extraction}b), defined for some module $m$ as:
\begin{equation}
    \centering
        \boldsymbol{P}^m \otimes \boldsymbol{M}_{m}^T =
            \begin{bmatrix}
            \boldsymbol{P}_{1m} \boldsymbol{M}_{m1} & \ldots & \boldsymbol{P}_{1m} \boldsymbol{M}_{m {D \choose 2}} \\
             \vdots & \ddots & \vdots \\
            \boldsymbol{P}_{\ell m} \boldsymbol{M}_{m1} & \ldots & \boldsymbol{P}_{\ell m} \boldsymbol{M}_{m {D \choose 2}}
            \end{bmatrix}
            ,
        % & \boldsymbol{P}_{1m} \boldsymbol{M}_{m2} 
        % \\ \boldsymbol{P}_{2m} \boldsymbol{M}_{m1} & \boldsymbol{P}_{2m} \boldsymbol{M}_{m2} & \ldots & \boldsymbol{P}_{2m} \boldsymbol{M}_{m {D \choose 2}} 
        %   \vdots &
        % & \boldsymbol{P}_{\ell m} \boldsymbol{M}_{m2} 
    \label{eq:module_outer_product}
\end{equation}
where the superscript $m$ denotes the column of the matrix $P$ (the presence vector of some module $m$) and $\otimes$ is the outer product. Intuitively, the reconstruction of all localities is related to matrix multiplication of $P$ and $M$, but instead of summing up entries across $M$, we take the maximum. In tensor terminology, outer products from Equation \ref{eq:module_outer_product} for each of the modules produce $M$ slices of size $\ell \times {D \choose 2}$, over which we take the per entry maximum to reconstruct $C$.

\textbf{AMF loss function.} We define the loss for locality $i$ as:
\begin{equation}
    \label{eq:amf_loss}
    \begin{split}
        \mathcal{L}(\boldsymbol{C}_i, &\boldsymbol{\hat{C}}_i) =\\ & =\frac{||\boldsymbol{C}_i||_2}{\sum_{j \neq i}^\ell ||\boldsymbol{C}_j||_2} \; d_{cos}\left(\mathbf{w}_i \odot \boldsymbol{C}_i, \mathbf{w}_i \odot \boldsymbol{\hat{C}}_i \right) \\
        & + \gamma \; \text{MAE}(\boldsymbol{C}_i, \boldsymbol{\hat{C}}_i),
    \end{split}
\end{equation}
where $||\,||_2$ are $L2$ norms, $\boldsymbol{\hat{C}}_i$ is the locality reconstruction (Equation \ref{eq:locality_module_formulation}), $\odot$ is the Hadamard product, $d_{\cos}$ is cosine distance defined as $d_{\cos}(\mathbf{x}, \mathbf{y}) = (1 - \cos(\mathbf{x}, \mathbf{y}))$, MAE is the mean absolute error with respective regularization hyperparameter $\gamma$. Overestimation mask $\mathbf{w}_i$ is defined as: %an overestimation mask, defined as:
\begin{equation}
    \centering
    w_{ic}= 
    \begin{cases}
        \nu,& \text{if } \boldsymbol{\hat{C}}_{ic} > \boldsymbol{C}_{ic} \\
        1,              & \text{otherwise}
    \end{cases}
    ,
    \label{eq:overestimation}
\end{equation}
where $w_{ic}$ is the weight for locality $i$ and correlation $c$, and  $\nu$ is an up-weighing hyperparameter (Appendix~\ref{app:module_extraction}). 
\begin{figure*}[!ht]
    \centering
    \includegraphics[width=\linewidth]{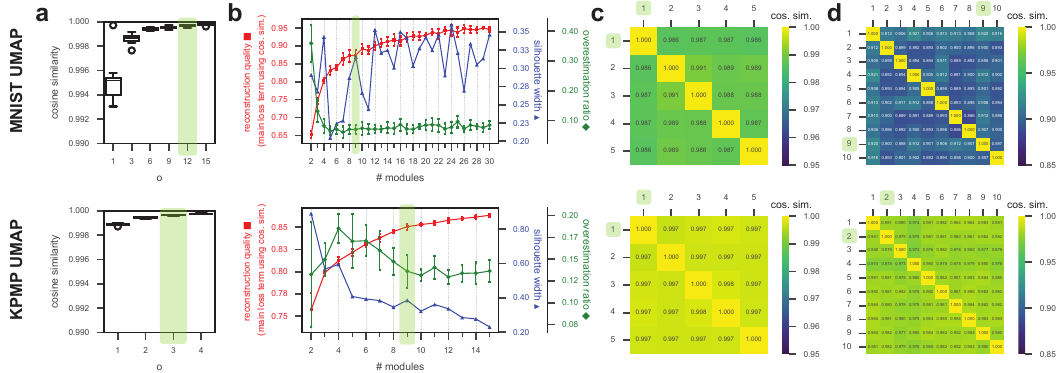}
    \caption{\textbf{LAVA stability experiments.} Top results for MNIST, bottom for KPMP UMAP embeddings. \textbf{a}, Cosine similarity of average correlations of locality placements when differing the overlap hyperparameter. \textbf{b}, Statistics of module extraction runs. \textbf{c-d}, Cosine similarity between presence-weighted averaged modules \textbf{c}, across different locality placements and \textbf{d}, within one locality placement.}
    \label{fig:stability}
\end{figure*}
The cosine distance is the main term of the loss function, whose semantics we deem appropriate for module extraction because we are more concerned with patterns of correlations than their absolute values. However, to prevent extraction of average, rather than reoccurring patterns, we up-weigh overestimation errors, similarly to quantile regression~\cite{koenker_quantile_2001} (Section \ref{sec:module_exp}). The MAE ensures modules are on the same scale as the locality correlations, despite the scale invariant cosine distance. The loss is optimized with ADAM~\cite{kingma_adam_2017} (Appendix~\ref{app:module_extraction}).

\textbf{Number of modules.} To decide the number of modules, we follow an approach used for non-negative matrix factorization (NMF)~\cite{alexandrov_deciphering_2013}, based on reconstruction and stability. For a range of values of $M$, we run AMF $R$ times using different random initializations of matrices $\boldsymbol{M}$ and $\boldsymbol{P}$.
To assess stability, we find $M$-medoids clusters using all $R \times M$ modules and calculate silhouette width~\cite{rousseeuw_silhouettes_1987} based on cosine similarity.
Higher stability indicates similar modules extracted across runs. Lower values of $M$ are desirable to guide the optimization towards more global shared patterns. However, too low a number might force subpatterns to merge, which should be reflected in poorer reconstruction. We thus look for an elbow of the loss across $M$ values, resolving ambiguity by favoring higher stability (Section \ref{sec:stability}).

% Once a number of modules is chosen, for the final modules, we select the run that has the lowest loss 
% ^ Removed because of the module complementarity argument, see experiments.

\section{Experiments}
\label{sec:experiments}
We perform several experiments to verify and showcase the utility of LAVA explanations. First, we evaluate the stepwise and end-to-end stability of the LAVA method. Second, we analyze whether LAVA explanations yield the properties expected by design. Third, we compare LAVA against the only existing alternative, the label-free explainability framework~\cite{crabbe_label-free_2022}. Finally, we use LAVA to explain UMAP-embedded kidney single-cell gene expression data, and reveal patterns related to kidney disease.
%\textcolor{red}{We perform several experiments to verify and utilize explanations of our method. We first perform a quantitative analysis of step-wise and end-to-end stability. We go on to highlight properties of LAVA modules through quantitative and qualitative experiments. We then also perform a comparison with an unsupervised feature importance method to illustrate essential differences of other methods in relation to our approach.} Finally, we use LAVA to explain a UMAP-embedded kidney single-cell gene expression dataset, revealing relevant biological patterns related to disease.

\textbf{Datasets.} Our experiments use two datasets: MNIST handwritten digit images~\cite{lecun_gradient-based_1998}, and KPMP\footnote[1]{Kidney Precision Medicine Project: https://www.kpmp.org/.} single-cell kidney gene expression~\cite{lake_atlas_2023}. The KPMP dataset measures the activity of $\sim$30k genes (features) for $\sim$225k kidney cells (samples)~\cite{stark_rna_2019}. We remove mitochondrial and ribosomal genes, and use only the top $1000$ most variable genes, following single-cell analysis guidelines~\cite{heumos_best_2023} (Appendix~\ref{app:datasets_embeddings_preprocessing}).

\textbf{Experimental setup.} 
We obtain two types of embeddings for MNIST, using UMAP manifold learning with default hyperparameters~\cite{mcinnes_umap_2020}) or a denoising convolutional autoencoder~\cite{crabbe_label-free_2022}. For KPMP, we use the publicly available UMAP embeddings.
%Firstly, we use the UMAP manifold learning method~\cite{mcinnes_umap_2020}: for MNIST, we generate a UMAP embedding using default hyperparameters, while for the KPMP data we use the UMAP embeddings provided with the dataset. Secondly, for the MNIST data, we also generate embeddings using a denoising convolutional autoencoder~\cite{crabbe_label-free_2022}.
% , allowing for comparison with work by \citet{crabbe_label-free_2022}. 
We apply LAVA to each set of embeddings, with neighborhood size~$n$ of $3000$ for MNIST and $500$ for KPMP (Appendix~\ref{app:locality_definition}), %For LAVA module extraction, we use 
and upweighing factor $\nu = 9$ for all experiments (Appendix~\ref{app:module_extraction}). %For modules of KPMP embeddings, we remove mitochondrial and ribosomal genes and use only the top $1000$ most variable genes following standard single-cell analysis guidelines~\cite{heumos_best_2023} (Appendix~\ref{app:datasets_embeddings_preprocessing}). 
To visualize MNIST results, we depict correlations of localities and modules as image-sized heatmaps, with pixel color denoting sum of correlations, and with blue lines between correlated pixels (Figure~\ref{fig:method_overview}, Appendix~\ref{app:mnist_visualization}).

\begin{figure*}[ht]
    \centering
    \includegraphics[width=\linewidth]{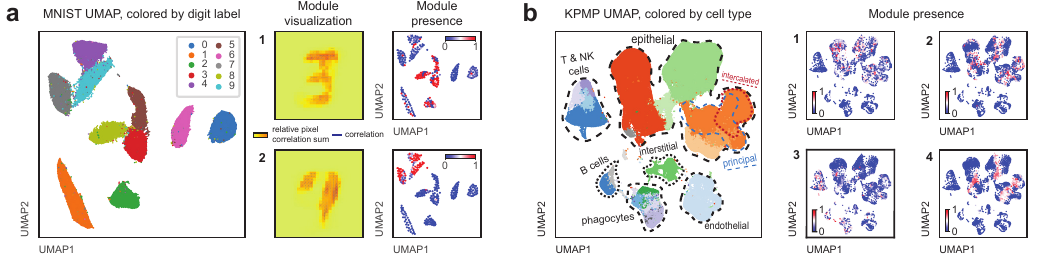}
    \caption{\textbf{LAVA explanations.} Labeled UMAP embeddings and LAVA explanations (modules \& presences) for \textbf{a}, MNIST  and \textbf{b}, KPMP.}
    \label{fig:intra_inter}
\end{figure*}

\subsection{Stability of LAVA Explanations}
\label{sec:stability}
%With this experiment, 
We assess stepwise and end-to-end solutions of LAVA to verify its ability to deliver stable explanations. %across different runs of locality placement and module extraction. 
\textbf{Setup.} Applying LAVA to UMAP embeddings of MNIST and KPMP, we assess stability of locality placement, module extraction, and end-to-end explanations as follows. %To assess 
\textbf{\textit{Locality placement}}: we run locality placement and representation (steps~1-2) $R_{L}=5$ times, for different values of overlap $o$; we then calculate $D \choose 2$ feature-pair average correlations across localities, and compare those between runs per $o$ value using cosine similarity (Figure~\ref{fig:stability}a). 
%We first run locality placement and calculate correlations $R_{LP}=5$ times, for different values of hyperparameter $o$. 
%To verify 
\textbf{\textit{AMF module extraction}}: using an arbitrarily chosen locality placement with $o=12$ for MNIST and $o=3$ for KPMP, whose $o$ values yield a compromise between fewer localities and higher stability, we run AMF module extraction (step~3) $R_M=10$ times for different numbers of modules~$M$; we report reconstruction quality (main loss term as cosine similarity instead of distance), overestimation ratio, and silhouette width of AMF runs per $M$ (Figure \ref{fig:stability}b). 
%Then, using a single locality placement for each of the datasets, we run module extraction $R_M=10$ times with different numbers of modules $M$ for each dataset, reporting the main loss term (as cosine similarity instead of distance), overestimation ratios, and silhouette width scores. 
%To assess 
\textbf{\textit{End-to-end LAVA explanations}}, or combined locality placement \& AMF module extraction: we run AMF for $M=9$ modules (criteria for $M$: Section~\ref{sec:module_extraction}) $R_{M}=10$ times for each of the $R_{L}=5$ locality placements; 
%We additionally run module extraction for the chosen number of modules $M=9$ (see below) on the $R_L=5$ different locality placements to verify end-to-end stability.
to assess similarity of the patterns underlying the modules irrespective of module organization (beyond silhouette width) across locality placements, we take a presence-weighted average of the module correlations per AMF run, average across runs per placement, and compare using cosine similarity (Figure~\ref{fig:stability}c); for an arbitrarily selected placement (Figure~\ref{fig:stability}c, highlighted), we also compare between presence-weighted module averages of individual AMF runs (Figure~\ref{fig:stability}d). \\

\textbf{LAVA locality placement is stable.}
%For the locality placement runs, we calculate average correlations of all localities for each run. We then calculate cosine similarities between the averages and report in Figure \ref{fig:stability}a. 
%We observe 
Feature-pair correlations are highly similar across locality placements %are highly similar 
for both MNIST and KPMP (cosine similarity $>0.993$), even with smaller numbers of localities influenced by overlap $o$, %and become increasingly similar with larger overlap $o$, 
indicating that locality placement is stable (Figure~\ref{fig:stability}a, Appendix \ref{app:loc_plac_stab}). %(e.g., higher than $0.993$ for $o=1$), becoming increasingly stable as we increase $o$ (Figure~\ref{fig:stability}a, Appendix \ref{app:loc_plac_stab}).
%
%\textbf{Properties of module extraction.}
\textbf{LAVA module extraction is stable.}
%We choose $o=12$ and $o=3$ for the MNIST and KPMP datasets, respectively, as a compromise between a lower number of localities and high stability. We then run module extraction (AMF) for a single locality placement run for both datasets and report in Figure \ref{fig:stability}b. 
As expected, larger numbers of modules $M$ lead to better reconstruction of the underlying correlations (Figure~\ref{fig:stability}b, red). 
%As expected, the loss term cosine similarity increases with larger numbers of modules, resulting in better reconstruction of the underlying correlations. 
For MNIST, we find relatively better reconstruction with small standard deviations (Figure~\ref{fig:stability}b, red), but also larger instability denoted by smaller silhouette widths (Figure~\ref{fig:stability}b, blue), pointing to successful optimization with multiple valid solutions. Here we note that silhouette width focuses on the goal of identifying a similar modular organization, which goes beyond capturing similar underlying patterns and could be too strict an objective for complex data. We therefore also evaluate pattern similarity below, and in Appendix \ref{app:mnist_exp}. 
%For MNIST, we find relatively lower silhouette widths irrespective of number of modules $M$, and smaller standard deviations of the loss, pointing to successful optimization but also a larger number of equally valid solutions (see below and in Appendix \ref{app:mnist_exp}).
For KPMP, a decrease in silhouette widths with increasing $M$ (Figure ~\ref{fig:stability}b, blue) indicates that smaller numbers of modules $M$ result in higher stability. Just like for MNIST, similar trends in reconstruction and stability for KPMP suggest the existence of multiple valid solutions. % using more stable solutions for smaller numbers, but similar instability with larger numbers. 
%\textcolor{red}{We additionally compare AMF modules to NMF components in Appendix \ref{app:nmf_amf}.} \\
%
%\textbf{Module correlations are stable across locality placements.} 
%\tetxbf{Module extraction is also stable across locality placements.}
%\textbf{Combined locality placement \& module extraction is stable.}
\textbf{LAVA delivers end-to-end stable explanations.}
%We decide to use $M=9$ modules for both datasets (criteria in Section \ref{sec:module_extraction}).
%For $M=9$, we run $R_{M}=10$ module extraction runs for each of the $R_{LP}=5$ different locality placements. For each run, we take a presence-weighted average of the module correlations, and then average across the runs for each locality placement. We compare using cosine similarity (Figure \ref{fig:stability}c). 
The LAVA modules capture highly similar correlation patterns across the $R_L=5$ locality placements (Figure \ref{fig:stability}c), and across AMF runs of a locality placement (Figure \ref{fig:stability}d). This indicates that LAVA explanations are highly stable end-to-end, throughout steps 1 to 3 of the LAVA method. 
%The average correlation patterns the modules capture are highly conserved across the $5$ different locality placements, indicating module extraction explains highly similar patterns even across different locality placements. 
%
%\textbf{Different module runs are complementary.}
%\textbf{LAVA explanations are stable.}
%For the selected modules from Figure \ref{fig:stability}b and c (highlighted green), we also calculate cosine similarities between the presence-weighted module averages of individual module extraction runs (Figure \ref{fig:stability}d). 
%We observe high cosine similarity between all runs. 
Together, high reconstruction quality with small standard deviations and highly similar patterns (Figures \ref{fig:stability}b-\ref{fig:stability}d), show that module runs capture very similar underlying patterns. We therefore conclude that the module runs represent differently factorized but equally valid solutions. 
%Together with low standard deviations between the loss functions and the highly similar patterns the different modules extraction runs capture (Figures \ref{fig:stability}b-\ref{fig:stability}d), we conclude the observed differences between module runs are not caused by algorithmic instability, but due to several equally valid, and can thus complementarily, solutions. 
%\textcolor{red}{As final explanations, we choose modules with the lowest loss (green highlights in Figure \ref{fig:stability}d), and leave for future work to examine whether specific use-cases can justify a different loss or combining multiple module runs to reduce overspecification.}

\begin{figure*}[ht]
    \centering
    \includegraphics[width=\linewidth]{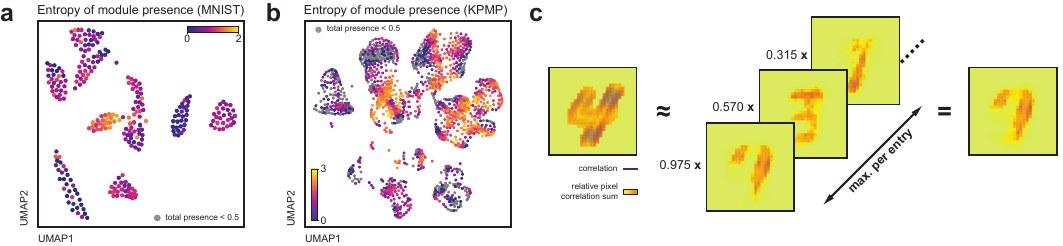}
    \caption{\textbf{LAVA module co-presences and locality reconstruction.} %\textbf{a-b}, 
    Co-presence as entropy of module presences per locality for UMAP embeddings of \textbf{a}, MNIST %(\textbf{a}) 
    and \textbf{b}, KPMP. %(\textbf{b}) UMAP embeddings, and an MNIST example of how modules reconstruct localities (\textbf{c}).
    \textbf{c}, Reconstruction of an example MNIST locality using its co-present modules.}
    \label{fig:overlap}
\end{figure*}

%\subsection{Module Properties}
\subsection{Properties of LAVA Explanations}
\label{sec:module_exp}
We investigate if the explanations extracted by LAVA possess the properties sought after by design. %Here, we qualitatively and quantitatively describe properties of LAVA modules.
\textbf{Setup.} We examine the LAVA modules obtained for UMAP embeddings of MNIST and KPMP as described in Section~\ref{sec:stability}.
%We examine the chose LAVA modules of the MNIST and KPMP UMAP embeddings from the above section.

\textbf{LAVA modules reveal globally reoccurring local subpatterns.}
For both datasets, the extracted modules represent globally reoccurring subpatterns not tied to neighboring localities (Figure~\ref{fig:intra_inter}). For instance, for MNIST, the two modules with the highest sum of presence in the latent space %(ranked by sum of presences in the latent space) 
capture correlation patterns shared between digits $3$, $5$, and $8$, and between digits $4$, $7$, and $9$ (Figure \ref{fig:intra_inter}a). %Visualizations of the two modules also show how the extracted correlations come from pixels shared across the digits these localities represent, as we would expect. 
We observe similar trends for KPMP embeddings (Figure \ref{fig:intra_inter}b), with some modules present across most major cell types (module $3$), and others pertaining to more local regions of specific cell types (e.g. epithelial cells for modules $1$, $2$, and $4$).
The module extraction uses no information about the spatial organization of localities, allowing it to identify subpatterns shared across or varying within clusters. %shared subpatterns across clusters and differentiate them within clusters.
\textbf{LAVA modules identify reoccurring, not averaged subpatterns.}
Modules exhibit overestimation error ratios below 15\% %of around $10\%$ or slightly higher 
(Figure \ref{fig:stability}b). Errors made up of predominantly underestimation indicate that locality-averaged patterns are avoided in favor of sparser but highly present patterns, much like in quantile regression~\cite{koenker_quantile_2001}.  Combined with the semantics of cosine similarity, the presence of a module in two localities implies not only a shared correlation subpattern, but also that such a subpattern can directly be observed in each of the localities (Figure~\ref{fig:intra_inter}, Appendices~\ref{app:mnist_exp}-\ref{app:nmf_amf}). 
%
%Comparison AMF vs. NMF
We further compare LAVA modules, extracted using AMF, to NMF components (Appendix~\ref{app:nmf_amf}). Without control for overestimation, NMF yields ratios of~50\%, indicating that NMF components capture average patterns across smaller sets of localities rather than finding more globally reoccurring subpatterns like AMF. Moreover, NMF relies on weighted sums of components, which is incoherent for correlations. %and makes them uninterpretable. 
%We further compare AMF modules to NMF components in Appendix \ref{app:nmf_amf}. With no control of the amount of overestimation, NMF components consistently capture comparatively more localized patterns with overestimation error ratios of 50\%, indicating they average correlations across smaller sets of localities rather than finding globally reoccurring subpatterns. Put together with the fact that NMF reconstruction are weighted sums of components, which is incoherent for correlations, NMF components are an uninterpretable and inaccurate representation of the correlations they are supposed to explain. %pattern of correlation, but also that the subpattern can directly be observed in both localities (see also appendices \ref{app:mnist_exp} and \ref{app:nmf_amf}).
\textbf{Multiple LAVA modules can coexist in a locality.} %Secondly, we show that multiple modules can be present at one locality. To verify that this happens
To investigate co-presence of LAVA modules, we scale their presences to sum to $1$ per locality and calculate Shannon entropy. %for each locality to sum to $1$, and calculate Shannon entropy to quantify module co-presence per locality. %in each dataset. 
Ignoring localities with non-scaled presence sums $<0.5$ to prevent inflated entropies from scaling, we get an average locality entropy of $0.68$ ($\pm 0.41$) for MNIST and $1.41$ ($\pm 0.75$) for KPMP (Figures \ref{fig:overlap}a-\ref{fig:overlap}b). For reference, $0$ means no module co-presence and $3.17$ corresponds to equal presence of all $9$ modules. %being equally present. 
The obtained entropies indicate module co-presences in both datasets, and more so in KPMP (Figures~\ref{fig:overlap}a-\ref{fig:overlap}b; Figure~\ref{fig:overlap}c %for an example of 
co-presence in a locality). %how modules overlap to reconstruct  correlations of a locality in MNIST. 
Allowing co-presence, a locality can be related to different subsets of localities via presences of modules, which may represent partially overlapping subpatterns of correlation. 

\begin{figure}[ht]
    \centering
    \includegraphics[width=\linewidth]{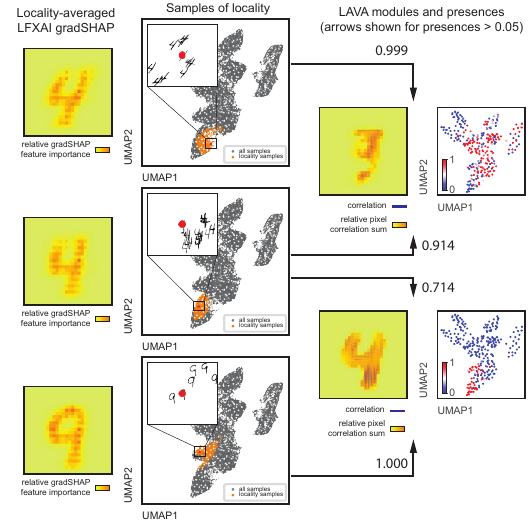}
    \caption{\textbf{LAVA \& label-free XAI.} Example (left) LFXAI feature importances  and (right) LAVA explanations -- modules and presences -- for localities of MNIST autoencoder embeddings. %of MNIST.
    }
    \label{fig:lfxai_comparison}
\end{figure}

%\subsection{LAVA vs. Unsupervised Feature Importance}
\subsection{Comparing LAVA and Label-Free Explainability}
\label{sec:lfxai_comp}
%In this experiment, 
We perform a qualitative comparison between LAVA and %feature importance based on 
the label-free explainability framework (LFXAI) by \citet{crabbe_label-free_2022}, to our knowledge the only other explainability method for unsupervised latent spaces. Comparison is challenging, since LAVA and LFXAI are fundamentally different. LAVA is natively designed to explain unsupervised models, while LFXAI uses adaptations of XAI for supervised ML. We focus on LFXAI-based feature importance, as it relates original features to embeddings, and is thus conceptually closer to LAVA than example importance. %at some level, whereas example importance offers a completely incomparable approach, relating samples in the latent space either based on similarity, or to the impact they have on the loss function.
In this context, LFXAI explanations are per sample linear combinations of feature importances derived independently for each latent dimension as model output. By contrast, LAVA captures feature covariation subpatterns across neighborhoods of sample embeddings in the multidimensional latent space. 
\textbf{Setup.} %Using the implementation of \citet{crabbe_label-free_2022}, 
We generate $4$-dimensional MNIST embeddings with a denoising convolutional autoencoder, using the implementation of \citet{crabbe_label-free_2022}. Note that LFXAI does not support embeddings without mapping function like UMAP. For LFXAI explanations, we use LFXAI with gradSHAP~\cite{lundberg_unified_2017} to generate feature importances for every sample. %First, we use their label-free explainability framework (LFXAI) to generate features importances using the gradSHAP~\cite{lundberg_unified_2017} implementation for all of the samples. 
For LAVA explanations, we run LAVA with number of neighbors $n=3000$, overlap $o=12$, and number of modules $M=8$ (Appendix \ref{app:additional_autoencoder}).
%We then run LAVA ($n=3000$, $o=12$), selecting $M=8$ modules for our explanations (Appendix \ref{app:additional_autoencoder}).
% Is there some stronger takeaway?
%\textbf{Conceptual differences.} 

\textbf{Explaining samples vs. embedding structure.} To enable comparison, we analyze explanations at the level of LAVA localities. We average LFXAI feature importances of the samples in each locality, and compare to LAVA modules for the same locality (Figure~\ref{fig:lfxai_comparison}, Appendix~\ref{app:additional_lfxai}). 
\begin{figure*}[th]
    \centering
    \includegraphics[width=\linewidth]{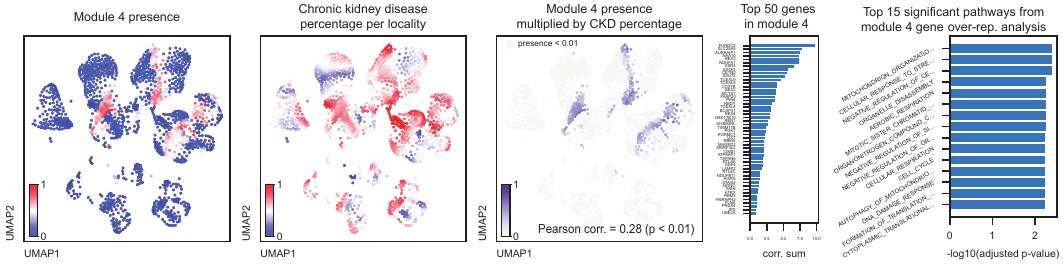}
    \caption{\textbf{LAVA module 4 for KPMP.} Module presence, chronic kidney disease (CKD) percentage, module-CKD overlap, gene analysis.}
    \label{fig:kpmp_experiment}
\end{figure*}
The LFXAI locality-averaged importances resemble the original digit images (Figure~\ref{fig:lfxai_comparison}, LFXAI), consistent with the purpose of revealing which features were relevant to place a sample in the latent space relative to a reference, empty image.  
%The feature importance scores tell us which pixels were used to place the sample in the latent space in relation to some reference, in this case an empty image (Appendix \ref{app:additional_lfxai}). 
%However, without significant and nontrivial adaption, it remains unclear 
However, it remains nontrivial if and how %to relate %or aggregate 
LFXAI explanations %from LFXAI 
can be exploited to %better 
understand the structure of the embeddings, %through the underlying data, 
regardless of which feature importance method is used. In contrast, LAVA explains how input features relate to the structure of the embeddings at a desired level of granularity. By capturing reoccurring feature subpatterns across localities, LAVA can further reveal relations between seemingly distant regions of the latent space that would not be apparent otherwise (Figure~\ref{fig:lfxai_comparison}, LAVA). 
%
%LAVA, on the other hand, cannot tell us which features were used to generate the latent embeddings, but uses the local structure of embeddings to capture globally reoccurring subpatterns of correlation across the latent space. %, and uncover relations between localities across the latent space. %between localities. %allows us to relate localities to one another based on these patterns.
%\textbf{LAVA is more suited for discovery than existing methods.} 
\textbf{LAVA enables discovery beyond explanation.} 
The LAVA and LFXAI methods provide complementary explanations for latent embeddings, each %offering meaningful information 
from a different perspective. Since one key goal of unsupervised learning is to identify patterns leading to new discoveries, the ability of LAVA to uncover hidden structure %across the latent space 
makes it ideally positioned to drive knowledge discovery beyond explainability (Figures~\ref{fig:intra_inter} and~\ref{fig:lfxai_comparison}). Importantly, when relating features to embedding structure, LAVA detects feature covariation irrespective of feature usage by the model. This allows LAVA to include features 
otherwise ignored, 
%seem less important to
for instance for exhibiting strong dependencies with others. As a result, LAVA delivers more complete explanations that address a common limitation of explainability methods~\cite{molnar2022}, and circumvent the Rashomon effect~\cite{breiman_statistical_2001, rudin_amazing_2024}. 
%Rather than extracting features used by the model to create the embeddings as is the case with feature importance, LAVA's modules reveal common reoccurring subpatterns of feature correlation implied by the modeling. In other words, if another model produced the same embeddings using a different subset of features, the feature importance would change but LAVA's explanations would not. This means that LAVA, unlike LFXAI, is able to detect complete sets of features that vary together locally, even if only a subset of those features was actually utilized by the model to produce the embeddings. This circumvents the Rashomon effect~\cite{breiman_statistical_2001, rudin_amazing_2024} and makes LAVA more suited for learning about the data through the model than existing methods.

\subsection{Explaining gene expression embeddings} 
\label{sec:kpmp_exp}
Finally, we enquire %investigate 
if LAVA %modules from Section \ref{sec:stability} 
can reveal biologically relevant patterns %within the 
in UMAP embeddings of KPMP data. 
\textbf{Setup.} We analyze LAVA modules for KPMP (Section~\ref{sec:stability}), together with %The KPMP dataset includes 
metadata on the human donors of the cells: we look at disease state, which is either healthy, acute kidney failure (ACF), or chronic kidney disease (CKD). To quantify the association between modules and a given disease state, %We first calculate percentages of samples from donor disease state per locality. We then 
we calculate Pearson correlation and respective p-value between module presence and percentage of samples labeled with the disease state (ignoring localities where a module is not present). % to gauge the association between module and disease state.
We further investigate if genes in those modules are involved in biological pathways related to the diseased state (Appendix~\ref{app:kpmp_exp}). 

\textbf{Module correlates with diseased samples.} Cells from donors with different disease states are spread %throughout 
across the latent space, with no apparent clustering. However, presence of module $4$ %(Figure~\ref{fig:kpmp_experiment}) 
%~\ref{fig:intra_inter}b) 
is positively correlated with the percentage of diseased samples ($0.23$, $p < 0.01$), and specifically CKD %samples specifically 
(Figure~\ref{fig:kpmp_experiment}, $0.28$, $p < 0.01$). Together with the fact that module $4$ is present in epithelial cells and not blood cells (Figures~\ref{fig:intra_inter}b and~\ref{fig:kpmp_experiment}), this suggests that %the correlation subpattern of 
module $4$ could be biologically relevant for kidney disease and CKD-related pathways. \textbf{Module genes and disease related pathways.} Overrepresentation analysis of biological pathways~\cite{fang_gseapy} for the %set of 
genes in module $4$ %correlations 
identifies $21$ significant pathways ($p<0.01$). To highlight biological relevance, we single out the %more general 
``cellular response to stress'' and ``DNA damage response'' pathways, linked to CKD in previous studies~\cite{patera_kidney_2024, schupp_dna_2016, molitoris_dna_nodate, wang_roles_2023}. Overall, our findings suggest that the local organization of KPMP latent UMAP embeddings across kidney %tissue 
cell types exhibits patterns of gene correlation related to kidney disease. This local structure %of sample organization 
is not apparent in the clustering of the data, but is notably discoverable through the LAVA modules. 

\section{Conclusions}
In this work, we frame the problem of explainability of latent embeddings as enabling reasoning about the latent spatial organization through the underlying input data. To that effect, we propose the LAVA method, which generates explanations in the form of modules, capturing local subpatterns of feature correlation reoccurring globally throughout the latent space. %explains input feature covariation in latent embeddings through modules, globally reoccurring local subpatterns of correlation found across the latent embeddings. 
We show that LAVA produces stable explanations at a desired level of granularity, complementary to existing label-free explainability approaches. In addition, LAVA uncovers hidden structure not apparent otherwise, which makes it especially suited for driving knowledge discovery. Specifically, we show that LAVA reveals domain-relevant patterns, such as visual parts of images and disease signals in cellular processes. 
For future work, we envision further exploration of spatially-guided reasoning about latent embeddings, especially for higher-dimensional latent spaces and supervised learning. %in which the spatial relations between embedded samples can be put into context and differentiated from one another based on specific targets of interest, building on top of existing feature and sample importance methods. 
%We see spatially-guided explainability of latent spaces having potential to extract more insight from ML models than existing methods, further aiding in model development and knowledge discovery.

\section*{Acknowledgments}

\textbf{The authors.}

Authors received funding from the US National Institutes of Health [grant numbers U54EY032442, U54DK134302, U01CA294527, and R01AI138581 to JPG; grant numbers U01DK133766 and R01AG078803 to IS and JPG]. Funders were not involved in the research, authors are solely responsible for this work.

\textbf{KPMP and CBR.}

The results here are in whole or part based upon data generated by the Kidney Precision Medicine Project. Accessed March 1, 2025. https://www.kpmp.org.

The Kidney Precision Medicine Project (KPMP) is supported by the National Institute of Diabetes and Digestive and Kidney Diseases (NIDDK) through the following grants: U01DK133081, U01DK133091, U01DK133092, U01DK133093, U01DK133095, U01DK133097, U01DK114866, U01DK114908, U01DK133090, U01DK133113, U01DK133766, U01DK133768, U01DK114907, U01DK114920, U01DK114923, U01DK114933, U24DK114886, UH3DK114926, UH3DK114861, UH3DK114915, and UH3DK114937. We gratefully acknowledge the essential contributions of our patient participants and the support of the American public through their tax dollars.

The authors acknowledge the University of Michigan Medical School Central
Biorepository (RRID:SCR\_026845) for providing biospecimen storage, management,
and distribution services in support of the research reported in this publication/grant application/presentation.

\section*{Impact Statement}
This paper presents work whose goal is to advance the field of Machine Learning. There are many potential societal consequences of our work, none which we feel must be specifically highlighted here.

\bibliography{example_paper}
\bibliographystyle{icml2026}

%%%%%%%%%%%%%%%%%%%%%%%%%%%%%%%%%%%%%%%%%%%%%%%%%%%%%%%%%%%%%%%%%%%%%%%%%%%%%%%
%%%%%%%%%%%%%%%%%%%%%%%%%%%%%%%%%%%%%%%%%%%%%%%%%%%%%%%%%%%%%%%%%%%%%%%%%%%%%%%
% APPENDIX
%%%%%%%%%%%%%%%%%%%%%%%%%%%%%%%%%%%%%%%%%%%%%%%%%%%%%%%%%%%%%%%%%%%%%%%%%%%%%%%
%%%%%%%%%%%%%%%%%%%%%%%%%%%%%%%%%%%%%%%%%%%%%%%%%%%%%%%%%%%%%%%%%%%%%%%%%%%%%%%
\newpage
\appendix
\onecolumn

\pagenumbering{roman}
\setcounter{page}{1}

% Make tables and figures and algorithms have prefixes and restart their counter
\setcounter{figure}{0}
\setcounter{table}{0} 
\setcounter{equation}{0} 
\setcounter{algorithm}{0} 
\renewcommand{\thefigure}{S\arabic{figure}}
\renewcommand{\thetable}{S\arabic{table}}  
\renewcommand{\theequation}{S\arabic{equation}}  
\renewcommand{\thealgorithm}{S\arabic{algorithm}}

\section{Methodology}
\label{app:methodology}

In this appendix we expand on the methodology of our method.

\subsection{Locality Definition}
\label{app:locality_definition}

\subsubsection{Defining the Size of Localities}
\label{app:hyperparameter_k}
The choice of hyperparameter $n$ is important as changing the size of localities can have a substantial influence on correlations that LAVA will calculate in step (2), and subsequently for module extraction in step (3). A data-driven way of choosing the value $n$ can be looking at which size of latent neighborhood most resembles the neighborhoods in the original feature size. In choosing such an $n$, we attempt to capture localities at a scale that most accurately represents organization in the original feature space. Another, more targeted, approach is motivating the choice of $n$ using metadata of interest. This might ground the scale of localities to existing knowledge and make findings more relevant for a specific downstream task. We employ both approaches in our experiments in Section \ref{sec:experiments}, and deliberate further in \ref{app:exp_hyperparameters}.

% As a specific example, in trajectory inference, pseudo-time~\cite{saelens_comparison_2019} might be used to determine a relevant scale at which to observe changes in variable associations. 

\subsubsection{Optimization of Locality Placement}
\label{app:opt_loc_plac}
For locality placement, we minimize the difference between the relative in-degree centralities of samples when we look at the neighborhoods of all samples, and the neighborhoods of localities (see Section \ref{sec:methods} and Equation \ref{eq:direct_loss}). For example, if some sample is in the $n$-nearest neighborhoods of $\%1$ of the samples in the embedding space, we want our localities to be placed so that the sample is a part of $\%1$ of the localities, i.e., so that the sample retains its in-degree centrality. We do this by optimizing parameters $\alpha$ and $\beta$ from Equation \ref{eq:direct_loss}, utilizing the $k$-means (here referred to as $\ell$-means) and DIRECT algorithms~\cite{jones_lipschitzian_1993}.

We use Manhattan instead of Euclidean distances in the loss function to focus less on outliers and approximate centrality more equally across the samples (equivalent to the difference between minimizing the mean absolute and mean squared errors). The rationale for using the weighing scheme in Equation \ref{eq:weighing_scheme} is forcing the computationally cheap $\ell$-means algorithm to optimize our locality placement, i.e. minimize Equation \ref{eq:direct_loss}. We want to weigh samples based on their centrality, since the $\ell$-means algorithm would otherwise minimize Euclidean distances rather than our objective. Furthermore, the weights in Equation \ref{eq:weighing_scheme} factor in the inverse average distances to offset the fact that $\ell$-means optimizes the sum of squared Euclidean distances between points, which means that lower density regions of the latent space might still have more probes (and localities), even when we weigh samples according to their centrality. Put together, we let the DIRECT algorithm optimize the weights of samples for the $\ell$-means algorithm, such that the localities produced by the $\ell$-means algorithm minimize our centrality-based loss function. Note that we use only the resulting centroids of the optimization as probes, while cluster assignments produced by the $\ell$-means algorithm are left unused. The $\ell$ localities are then simply the $n$-sized neighborhoods around each of the $\ell$ probes. Additionally, the DIRECT algorithm requires a range of values for the parameters in which the loss function is assumed to be convex. We fix this to $[-2D_L, 2D_L]$ for both $\alpha$ and $\beta$, where $D_L$ is the dimensionality of the latent space. We conjecture that, at least approximately (since $\ell$-means is stochastic, the loss function is convex in that interval and contains a good solution). Loosely, as mentioned above, we want the weighing scheme to be able to overpower the overrepresentation of low-density regions that might occur if we would simply weigh samples based on their centrality. We argue about it in terms of hyperball volume: the hyperball volume is proportional to the $D_L$-th power of its radius, which for a neighborhood of a sample we can approximate by the average distance to its neighbors, where $D_L$ is the dimensionality of the latent space. To battle overrepresentation, we then want to weigh samples with larges ``volumes'' (samples in low density regions) inversely to their volume, so as to effectively force $\ell$-means to give them less importance than they might otherwise have given the squared Euclidean distance optimization. In the two-dimensional case, the hyperball volume is simply the area of the circle, which is proportional to the square of the radius (the average distance of the $n$ nearest neighbors). To allow weighing in such a fashion, we allow $\alpha$ to be at least $2$, and we further multiply by $2$ to allow for exaggeration of this principle. Since we cannot be sure how this weighing will interact with centrality weighing, we allow both parameters to also be negative, and give them the same range, arbitrarily.  In our limited experimentation, and with no ground truth to compare to, this seemed to yield well-placed localities results (Figures \ref{fig:mnist_probes} and \ref{fig:kpmp_probes}).

The pseudocode, ommiting details about DIRECT, is shown in Algorithm \ref{alg:direct_optimization}. We use the scipy\footnote{\hyperref[https://scipy.org/]{https://scipy.org/}.} implementation of the DIRECT algorithm, and the scikit-learn\footnote{\hyperref[https://scikit-learn.org/]{https://scikit-learn.org/}.} implementation of the $k$-means algorithm.

\begin{algorithm}[h]
    \caption{Locality Placement Algorithm.}
    \label{alg:direct_optimization}
    \begin{algorithmic}
        \STATE \textbf{require} {$\mathbf{X} \in \mathbb{R}^{E\times D_L}$%\mathcal{H}^E$
        , $\ell$}
        
        \STATE \textbf{initialize} $nearest\_neighbors[E, n]$, $in\_neighborhood[E]$, $avg\_n\_distance[E]$

        \STATE $nearest\_neighbors \leftarrow \text{calculate\_nearest\_neighbors}(\mathbf{X}, n)$ 
        \FOR{$i = 1$ {\bfseries to} $E$}
        \STATE $in\_neighborhood[i] \leftarrow \frac{1}{E}\sum_{j=1}^E{\mathbbm{1}\left[\mathbf{X}_i \in nearest\_neighbors[j]\right]}$
        \STATE $avg\_n\_distance[i] \leftarrow \frac{1}{n}\sum_{j=1}^n{||\mathbf{X}_i - nearest\_neighbors[i, j]||_2}$
        \ENDFOR

        \STATE \textbf{initialize} $localities[\ell, n]$, $probe\_set[\ell]$, %$\alpha_{range}[2]$, $\beta_{range}[2]$, 
        $\alpha_{opt}$, $\beta_{opt}$
        %\State $\alpha_{range} \leftarrow [-4, 4]$ \Comment{value ranges for parameters}
        %\State $\beta_{range} \leftarrow [-4, 4]$
        \STATE $\alpha_{opt}, \beta_{opt} \leftarrow \text{locality\_optimization$_{DIRECT}$\footnotemark}(\mathbf{X}, \ell, in\_neighborhood, avg\_n\_distance)$%, \alpha_{range}, \beta_{range})$
        
        \STATE $probe\_set \leftarrow \text{k\_means}(\ell, \alpha_{opt}, \beta_{opt}).\text{centroids()}$
        \STATE $localities \leftarrow \text{calculate\_nearest\_neighbors}(probe\_set, n)$
    \end{algorithmic}
\end{algorithm}
\footnotetext{More information in Section \ref{app:opt_loc_plac}.}

\subsubsection{Defining the Number of Localities}
To determine the number of localities (and respective probes), $\ell$, we look at three factors: the number of embeddings $E$, the size of the neighborhood $n$, and an additional hyperparameter $o$ relating to the overlap between localities. $\ell$ is then defined as:
\begin{equation}
    \centering
    \ell = [E * o / n],
\end{equation}
where $[\,]$ denotes rounding to the nearest integer. The hyperparameter $o$ determines the number of localities given a dataset and neighborhood size $n$. If all samples were of a similar centrality (e.g., if data are uniformly spread-out across some space), $o=x$ would mean that each sample would be represented by approximately $x$ localities. A larger $o$ will produce more overlap between localities and enable a representation that more closely approximates the original in-degree centralities of samples, but will be more expensive computationally and harder to manage in subsequent analysis. In Section \ref{sec:experiments} and Appendices \ref{app:loc_plac_stab}, we examine the stability of locality placement for UMAP embeddings of two different datasets in relation to hyperparameter $o$. Assuming stable (consistent) locality placement, we recommend using smaller values of $o$ in the range of $2$ to $20$, depending on the size of the dataset, so that the resulting number of localities $\ell$ is not above several thousand, but still captures the various correlation patterns and how they change across the latent space.

\subsection{Locality Representation}
\label{app:calculating_correlations}
Correlations of each locality are calculated using the standard formulation of the Spearman correlation:
\begin{equation}
    r_s(\mathbf{X}_i, \mathbf{X}_j) = \frac{cov\left[\text{R}(\mathbf{X}_i), \text{R}(\mathbf{X}_j)\right]}{\sigma_{\text{R}(\mathbf{X}_i)}, \sigma_{\text{R}(\mathbf{X}_j)}},
\end{equation}
where $X_i$ and $X_j$ are two original input features and $\text{R}$ is a ranking function. Covariance $cov$ and standard deviations $\sigma$ are estimated from the samples within a locality using standard estimators. Tied ranks are averaged.

Importantly, if one of the variables has the same value in $75\%$ of the neighborhood, that correlation is set to $0$ for that locality. This is done to prevent overestimation of correlations in cases where there is very little change. Notably, this has been the case for both the MNIST and KPMP datasets in our experiments, as the values of most pixels in MNIST are $0$, and gene expression at the single-cell level in KPMP (and more generally) is very sparse. The value of $75\%$ was set arbitrarily, to allow for some leniency, while avoiding meaningless correlations based on lower numbers of samples.

We also note that other measures of association, and even univariate measures of variation, could be used depending on the intended use case. However, the different locality representation might require significant alternations to the subsequent module extraction.

\subsection{Module Extraction}
\label{app:module_extraction}

\subsubsection{AMF loss function and hyperparameters}
The AMF loss function (Equation \ref{eq:amf_loss}) is discussed in Section \ref{sec:module_extraction}. Here, we provide some additional detail and motivation.

\textbf{Different metrics of (dis)similarity.} We decide to use cosine distance for the main term of our loss function. However, a different metric (or divergence) for assessing multivariate (dis)similarity might be equally suitable, e.g., the Kullback-Leibler divergence. We did not find strong argumentation about whether the semantics of one of these might be more appropriate for our task than the other. Work relating to semantics and efficiency of these metrics also suggests ``triangular distance'' as a more appropriate choice in some contexts~\cite{connor_tale_2016}, but is seldom used in contemporary literature.

\textbf{Up-weighing overestimation and connection to quantile regression.}
The $\nu$ hyperparameter in Equation \ref{eq:amf_loss} determines how up-weighted overestimation errors will be in our loss. Weighing the errors is done similarly to how it is done in univariate quantile regression. The loss for univariate $\tau$-th quantile regression, also known as the pinball loss~\cite{steinwart_estimating_2011}, is defined as:
\begin{equation}
    \mathcal{L}_{\tau}(y, f(x)) =
    \begin{cases}
        \tau \, |y - f(x)|& \text{if } y \geq f(x) \\
        (1 - \tau) \, |y - f(x)| & \text{if } y < f(x)
    \end{cases}
    .
    \label{eq:pinball}
\end{equation}
Practically speaking, univariate quantile regression produces a weighing scheme between overestimation and underestimation errors: if $\tau > 0.5$, we are less concerned with overestimating $y$ using $f(x)$, and vice versa. Using a $\tau = 0.5$ is equal to doing median regression, which corresponds to a regression that minimizes the mean absolute error rather than using the more standard mean squared error. For more intuition, we refer to previous work~\cite{koenker_quantile_2001, steinwart_estimating_2011}.
 
The idea for module extraction is to retain the semantics of similar, scale-invariant patterns we get with cosine distance or a similar metric, but allowing for reasoning about the reconstructed correlations in terms similar to quantile regression. Retaining the error weighing with a $\tau < 0.5$ is therefore done to prevent averaging across samples and extract only shared, reoccurring correlations (Section \ref{sec:module_extraction}). Quantile-regression-like reconstructions will allow us to say that correlations present in module-based reconstructions are actually present in the localities with some certainty, or, in terms of errors, that the value of a correlation will be overestimated by at most $\tau$ (approximately). We can quantify whether this is actually the case by looking at the overestimation errors (Figure~\ref{fig:stability}b), which we calculate as the amount of overestimation errors divided by the total absolute error.

Our hyperparameter $\nu$ that weighs overestimation effectively corresponds to the ratio of weights for the overestimation and underestimation:
\begin{equation}
    \centering
    \nu = \frac{1 - \tau}{\tau}
    .
\end{equation}
Using $\nu$ for weighing overestimation errors when using the mean absolute errors would be equivalent to doing $\tau$-th quantile regression. We have set the hyperparameter $\nu = 9$ in all of our experiments. If we look at the pinball equivalent $\tau$, this would correspond to $0.1$-th quantile regression. We cannot make direct parallels since we do not optimize the mean absolute error, but we can observe relative amounts of overestimation, which seem to more-or-less correspond to values we would expect if we would do quantile regression (Figure~\ref{fig:stability}b).

\textbf{Extra: Presence fine-tuning.} While we do not use it in our experiments, our implementation also includes what we call presence fine-tuning. After selecting modules, we can fix the module matrix $\mathbf{M}$ and optimize the presence matrix $\mathbf{P}$ with a pinball loss function (Equation \ref{eq:pinball}), to guarantee the property that each locality (independent when $\mathbf{M}$ is fixed) has an overestimation ratio exactly equal to a chosen $\tau$. Notice that the modules we extract are still contingent on the overestimation weighing used in the main loss term (because modules and presences are optimized jointly), which means that using weighing in the main loss term is not equivalent to first extracting the modules without any constraints and then fine-tuning at the end.

\textbf{Scale regularization.} The regularization term in Equation \ref{eq:amf_loss} is a simple mean absolute error term designed to retain scale of reconstructions. The idea behind scaling is simply to make reconstructions of the same scale as the actual locality correlations, as cosine distance is in itself scale invariant. While the cosine distance should remain relatively unaffected by this regularization, given a high enough hyperparameter $\gamma$, matching scale of the actual locality correlations and our reconstructions will allow us to observe the overestimation ratio of solutions (Section~\ref{sec:experiments}; Figure~\ref{fig:stability}b). Furthermore, due to the scale invariance of cosine distance, it is also imaginable that modules avoid overestimation simply by retaining low correlations values across the board: the scaling will push optimization away from such solutions. It is crucial that the scale regularization contains no overestimation weighing, meaning that the ratio of overestimation is purely a result of the up-weighing in the first term of the loss function, as this ensures the retainment of the semantics of reoccurring patterns, rather than focusing on absolute values of our solutions. 
We set $\gamma = 0.0001$ as the default value for scale regularization to have a low impact on the loss function. We also ended up manually setting this hyperparameter for the KPMP dataset to $\gamma = 10.0$ in the experiments (see Section \ref{app:hp_module_ex}), as we noticed $\gamma=0.001$ had no effect on optimization because of the low amounts of observed correlation. 

\textbf{Optimization hyperparameters.} The matrices $\mathbf{P}$ and $\mathbf{M}$ are optimized to minimize the loss function (Equation \ref{eq:amf_loss}) using gradient descent with minibatches of $64$ samples (arbitrarily set), the ADAM optimizer~\cite{kingma_adam_2017} with default hyperparameters, and clamping. Clamping is used to force any entries of matrices $\mathbf{P}$ and $\mathbf{M}$ to be within the $[0, 1]$ interval after each step in training. Training is performed until convergence, which we define as no improvement in the loss function above $1\%$ in $100$ epochs. After convergence, we used the matrices from the iteration with the lowest loss as the final solutions as described in Section~\ref{sec:experiments}. We developed the code using the PyTorch library\footnote{\hyperref[https://pytorch.org/]{https://pytorch.org/}.}.

\section{Experiments}
\label{app:exps}

In this appendix we provide additional information about our experiments and show additional results and figures.

% TODO: add GitHub for camera-ready
\subsection{Software and hardware}
All our code was developed using Python 3.8 and its libraries\footnote{Code included with supplementary material and will be made available online, conditional on acceptance.}. We ran experiments on Linux compute nodes, comprising of a number of different CPUs and GPUs, details of which we omit to ensure anonymity. Locality placement and correlation calculations took several hours for both datasets. A single module extraction run would take about $5$ to $30$ minutes, increasing with the number of modules, $M$. We also tested all of our code on an Apple M3 Pro with 12 CPU cores and 18GB of RAM, which is what we also used to run Jupyter notebooks containing result analyses and plot generation.

\subsection{Datasets, embeddings, and preprocessing}
\label{app:datasets_embeddings_preprocessing}

\subsubsection{MNIST}
The MNIST dataset was downloaded using the Scikit-learn Python library\footnote{\hyperref[https://scikit-learn.org]{https://scikit-learn.org}.} library. No preprocessing was performed on the data. Prior to running LAVA, we created: (a) a 2-dimensional UMAP embedding of the dataset using the \textsc{UMAP}\footnote{\hyperref[https://umap-learn.readthedocs.io]{https://umap-learn.readthedocs.io}.} library, with default hyperparameters; (b) a 4-dimensional autoencoder embedding of the dataset using the denoising autoencoder from the Label-Free Explainability paper~\cite{crabbe_label-free_2022}.

\subsubsection{KPMP single-cell kidney atlas, v1.5}

The KPMP single-cell kidney atlas was downloaded from the CELLxGENE website\footnote{\hyperref[https://cellxgene.cziscience.com]{https://cellxgene.cziscience.com}. Specific dataset version: \hyperref[https://datasets.cellxgene.cziscience.com/f5b6d620-76df-45c5-9524-e5631be0e44a.h5ad]{https://datasets.cellxgene.cziscience.com/f5b6d620-76df-45c5-9524-e5631be0e44a.h5ad}.}. The dataset contains a UMAP embedding of the data preprocessed using the Seurat library\footnote{\hyperref[https://satijalab.org/seurat/]{https://satijalab.org/seurat/}.}, which we used for our experiments. More information can be found at the KPMP data portal website\footnote{\hyperref[https://atlas.kpmp.org/]{https://atlas.kpmp.org/}.}. The UMAP of the data can also be explored on the CELLxGENE website\footnote{\hyperref[https://cellxgene.cziscience.com/e/dea717d4-7bc0-4e46-950f-fd7e1cc8df7d.cxg/]{https://cellxgene.cziscience.com/e/dea717d4-7bc0-4e46-950f-fd7e1cc8df7d.cxg/}.}. For module extraction, we removed mitochondiral, ribosomal, and mitochondrial ribosome genes. Then, we selected the top $1000$ most variable genes, based on the available gene expression data (data after normalization and log transformation).

\subsection{LAVA hyperparameters}
\label{app:exp_hyperparameters}

\subsubsection{Size of locality, $n$}
\label{app:exp_locality_size_k}

\begin{figure}[h]
    \centering
    \includegraphics[width=0.32\linewidth]{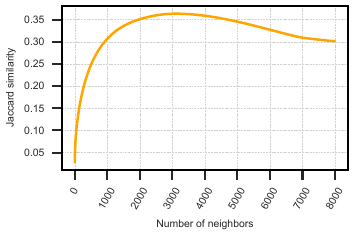} 
    \caption{Average Jaccard similarity between original and latent space neighborhoods, calculated for a UMAP of the MNIST dataset with default parameters. Despite the default neighborhood size parameter being $15$, we found the highest original neighbor retention to be around $3000$ neighbors.}
    \label{fig:umap_jaccard}
\end{figure}

\textbf{Data-driven approach (MNIST).} A data-driven way of choosing the value $n$ can be looking at the Jaccard similarity of neighborhoods of varying sizes in the original space and the latent space.  Even though methods such as UMAP have hyperparameters that represent the size of neighborhoods that will be optimized for, different neighborhood sizes might end up being optimized (depending on the choice of other hyperparameters and the dataset). For the UMAP of the MNIST dataset used in our experiments (Section \ref{sec:experiments}), we found that with the default hyperparameter of $15$ nearest neighbors, the resulting embedding neighborhoods were most representative of the neighborhoods in the original feature space when $k \approx 3000$ (Figure \ref{fig:umap_jaccard}), which was the value we used for LAVA's hyperparameter in our experiments with MNIST. For simplicity, we also retained $k=3000$ for the LAVA experiments with the autoencoder-generated embeddings.

\textbf{Meta-data approach (KPMP).} For the KPMP dataset, we were interested in observing patterns of gene correlation on a sub-cell type level. Looking at cell types in the data, we observed the median number of cells for each cell type was $3592$, while the smallest cell type had $148$ cells (Table \ref{tab:cell_type_sizes}). Accordingly, we chose neighborhoods of size $k=500$, as this seemed like a large enough number to capture correlations between gene expression that cannot be considered noise, while small enough to explore differences within most of the cell types.
\begin{table}[h]
    \centering
    \caption{Quartile statistics of the numbers of samples per cell type.}
    \begin{tabular}{|c||c|c|c|c|c|} \hline 
         &  min.&  25\%&  50\%&  75\%& max.\\ \hline \hline 
         \# cells&  148&  1665&  3592&  9105& 54025\\ \hline
    \end{tabular}
    \label{tab:cell_type_sizes}
\end{table}

\subsubsection{Module extraction hyperparameters}
\label{app:hp_module_ex}

\begin{figure}[h]
     \centering
     \begin{subfigure}[b]{0.45\textwidth}
         \centering
         \includegraphics[width=\textwidth]{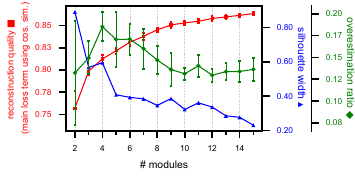}
         \caption{$\gamma=10.0$}
     \end{subfigure}
     \hfill
     \begin{subfigure}[b]{0.45\textwidth}
         \centering
         \includegraphics[width=\textwidth]{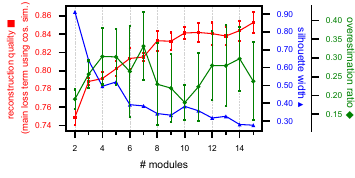}
         \caption{$\gamma=1.0$}
     \end{subfigure}
     \begin{subfigure}[b]{0.45\textwidth}
         \centering
         \includegraphics[width=\textwidth]{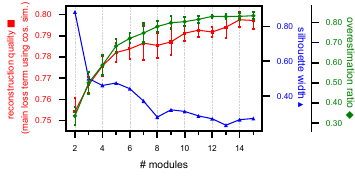}
         \caption{$\gamma=0.01$}
     \end{subfigure}
     \hfill
     \begin{subfigure}[b]{0.45\textwidth}
         \centering
         \includegraphics[width=\textwidth]{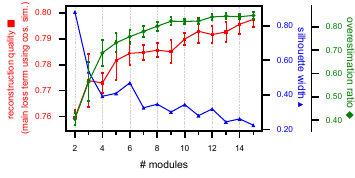}
         \caption{$\gamma=0.001$}
     \end{subfigure}
     \caption{Module extraction runs with different regularization hyperparameter $\gamma$.}
     \label{fig:kpmp_mod_ex_alpha}
\end{figure}

For module extraction on both MNIST and KPMP, we use default hyperparameters, as defined in Appendix \ref{app:module_extraction}, with one exception: the $\gamma$ regularization hyperparameter from Equation \ref{eq:amf_loss} in KPMP module extraction. In our runs, we noticed that using low values for $\gamma$ resulted in high overestimation error ratios for KPMP experiments. This pointed to the fact that regularization was mostly likely not affecting the loss function. We assumed this was due to the sparsity of gene expression, and subsequent low correlations in many localities. Running module extraction several times, using higher values of $\gamma$, confirmed this hypothesis (Figure \ref{fig:kpmp_mod_ex_alpha}). Increasing the strength of regularization seemed to have no or negative effect on the quality or stability of solutions (the main loss term and silhouette width, respectively). For our experiments, we ended up using $\gamma = 10.0$, resulting in relatively low and stable error overestimation ratios (Figure \ref{fig:kpmp_mod_ex_alpha}a).

\subsection{Visualizations of correlations for the MNIST dataset}
\label{app:mnist_visualization}

\begin{figure}[h]
    \centering
    \includegraphics[width=0.825\textwidth]{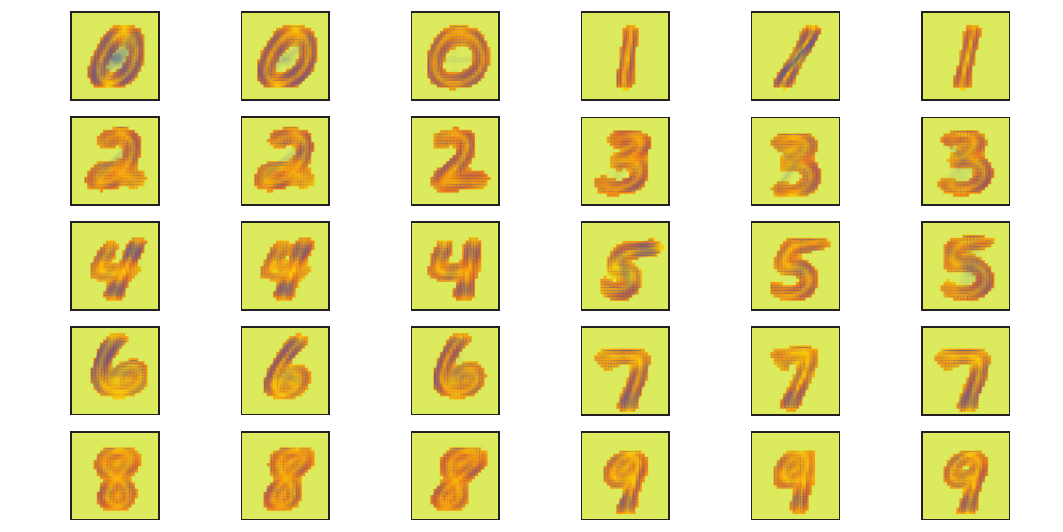}
    \caption{Example visualizations of correlations at various localities from our experiments with the MNIST dataset.}
    \label{fig:supp_mnist_visuals}
\end{figure}

The MNIST dataset consists of images that are $28$ pixels high and wide, totaling $784$ pixels. To show the importance of each feature in a correlation matrix, we create a heatmap of the same dimensionality ($28 \times 28$), where the value of a pixel correspond to the sum of the correlations that pixel participates in. Coloring is relative, meaning that it conveys relative importance of pixels for one locality or module (and its corresponding correlation matrix) only, and not across different visualizations. Furthermore, we add lines to connect pixels that are correlated to one another. This helps visualize the correlations which cause some pixels to be important in the first place. To avoid clutter, we exponentiate correlations to the power of $3$, and then remove correlations that are lower than $0.1$ after exponentiation (arbitrarily, for visual clarity). The widths of lines that represent correlations are proportional to their exponents to differentiate between stronger and weaker correlations. Some additional examples of visualizations can be seen in Figure \ref{fig:supp_mnist_visuals}.

Intuitively, we can think about these visualizations as a way to visualize pixel movement in a group of samples. In the case of local groups of similar images of digits, like we had in our experiments with MNIST, we can assume correlations will most likely reflect changes in pixel intensity due to minor translations, rotations, stretchings, or shrinkings of images. Firstly, for all of these transformations, we can expect that nearby pixels will be correlated, which we can see in all visualizations (Figure \ref{fig:supp_mnist_visuals}). For the latter three transformations, we can also expect further apart pixels to show patterns of correlation, which we see more clearly in correlations relating to the digits $0$ and $1$, as well as some other localities.

\clearpage

\subsection{Locality Placement}
\label{app:loc_plac_stab}
\begin{figure}[h]
     \centering
     \begin{subfigure}[b]{0.32\textwidth}
         \centering
         \includegraphics[width=\textwidth]{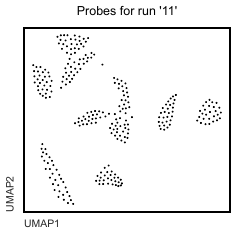}
         \caption{Locality (probe) placement 1.}
     \end{subfigure}
     \hfill
     \begin{subfigure}[b]{0.32\textwidth}
         \centering
         \includegraphics[width=\textwidth]{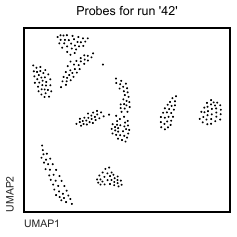}
         \caption{Locality (probe) placement 2.}
     \end{subfigure}
     \hfill
     \begin{subfigure}[b]{0.32\textwidth}
         \centering
         \includegraphics[width=\textwidth]{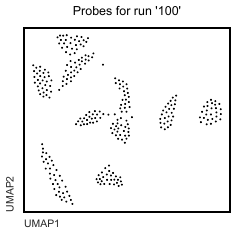}
         \caption{Locality (probe) placement 3.}
     \end{subfigure}
     \hfill
     \begin{subfigure}[b]{0.32\textwidth}
         \centering
         \includegraphics[width=\textwidth]{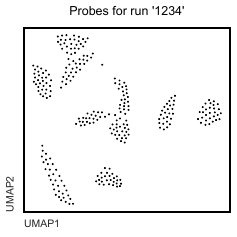}
         \caption{Locality (probe) placement 4.}
     \end{subfigure}
     \begin{subfigure}[b]{0.32\textwidth}
         \centering
         \includegraphics[width=\textwidth]{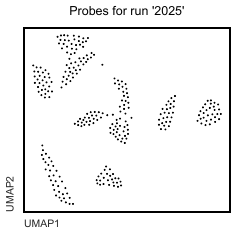}
         \caption{Locality (probe) placement 5.}
     \end{subfigure}
     \caption{The $5$ different probe placements for the UMAP MNIST embeddings, $o=12$.}
     \label{fig:probe_placements_mnist}
\end{figure}

\begin{figure}[h]
     \centering
     \begin{subfigure}[b]{0.32\textwidth}
         \centering
         \includegraphics[width=\textwidth]{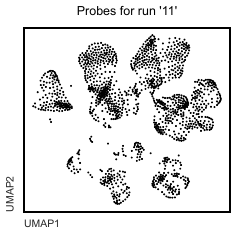}
         \caption{Locality (probe) placement 1.}
     \end{subfigure}
     \hfill
     \begin{subfigure}[b]{0.32\textwidth}
         \centering
         \includegraphics[width=\textwidth]{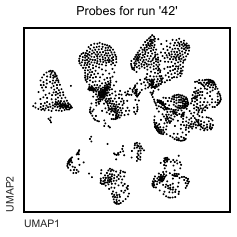}
         \caption{Locality (probe) placement 2.}
     \end{subfigure}
     \hfill
     \begin{subfigure}[b]{0.32\textwidth}
         \centering
         \includegraphics[width=\textwidth]{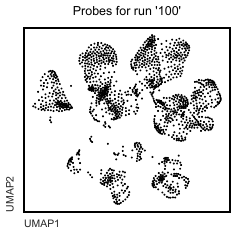}
         \caption{Locality (probe) placement 3.}
     \end{subfigure}
     \hfill
     \begin{subfigure}[b]{0.32\textwidth}
         \centering
         \includegraphics[width=\textwidth]{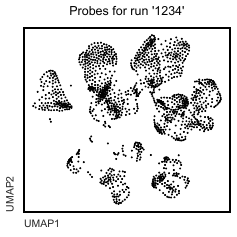}
         \caption{Locality (probe) placement 4.}
     \end{subfigure}
     \begin{subfigure}[b]{0.32\textwidth}
         \centering
         \includegraphics[width=\textwidth]{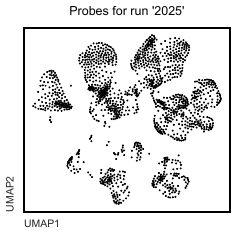}
         \caption{Locality (probe) placement 5.}
     \end{subfigure}
     \caption{The $5$ different probe placements for the UMAP KPMP embeddings, $o=3$.}
     \label{fig:probe_placements_kpmp}
\end{figure}

\begin{figure}[h]
     \centering
     \begin{subfigure}[b]{0.33\textwidth}
         \centering
         \includegraphics[width=\textwidth]{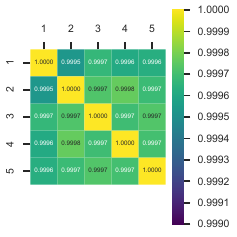}
         \caption{Cosine similarity between the average locality correlations for the MNIST UMAP embeddings, $o=12$.}
     \end{subfigure}
     \hspace{0.17\textwidth}
     \begin{subfigure}[b]{0.33\textwidth}
         \centering
         \includegraphics[width=\textwidth]{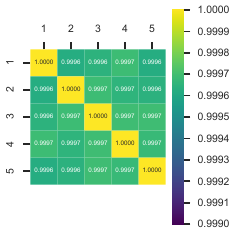}
         \caption{Cosine similarity between the average locality correlations for the KPMP UMAP embeddings, $o=12$.}
     \end{subfigure}
     \caption{Cosine similarities across the $5$ different locality placement runs for the MNIST and KPMP UMAP embeddings. Locality placements are named by the integers used as seeds for the different runs.}
     \label{fig:probe_placement_cos_sim_mat}
\end{figure}

Here, we show some additional results for experiments relating to locality placement. The experiments were done using the UMAP embeddings of the MNIST and KPMP dataset. In the main paper (Figure \ref{fig:stability}a), we show how similar the average locality correlations are across $5$ different locality placement runs while varying the hyperparameter $o$. For more qualitative comparison, here we also show at the specific probe placements for the chosen hyperparameters $o=12$ and $o=3$ (for MNIST and KPMP, respectively) in Figure \ref{fig:probe_placements_mnist} and Figure \ref{fig:probe_placements_kpmp}, respectively. While we do see minor differences, the number of probes across the different areas and their locations remain relatively similar. Furthermore, our results (Figure \ref{fig:stability}a, specifically the hyperparameters highlighted in green) indicate that the correlation patterns they capture are extremely similar. To further substantiate this claim, we also show cosine similarities of the average locality correlations across the 5 different runs in Figure \ref{fig:probe_placement_cos_sim_mat}.

\clearpage

\subsection{End-to-End Module Extraction Stability}
\label{app:end_to_end_mod_ex_stab}

Here, we show some additional results relating to module extraction stability, to reinforce the case that: (a) LAVA can produce consistent and stable results; (b) the instability of module extraction in our experiments is caused by the multiplicity of possible factorizations (overspecification), rather than algorithmic instability of our methodology. These results are based on LAVA runs on the UMAP embeddings of the MNIST and KPMP datasets from Section \ref{sec:experiments}. 

\begin{figure}[h]
     \centering
     \begin{subfigure}[b]{0.33\textwidth}
         \centering
         \includegraphics[width=\textwidth]{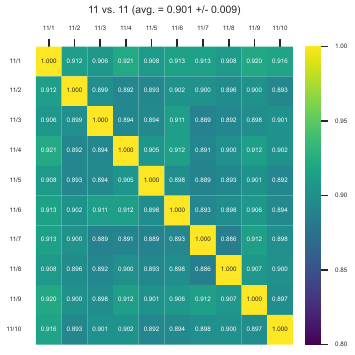}
         \caption{}
     \end{subfigure}
     \hfill
     \begin{subfigure}[b]{0.33\textwidth}
         \centering
         \includegraphics[width=\textwidth]{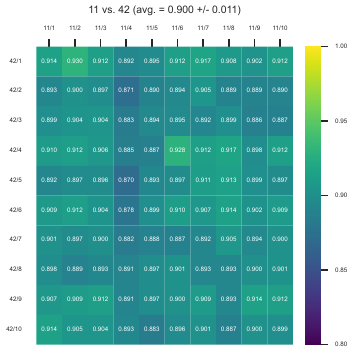}
         \caption{}
     \end{subfigure}
     \hfill
     \begin{subfigure}[b]{0.33\textwidth}
         \centering
         \includegraphics[width=\textwidth]{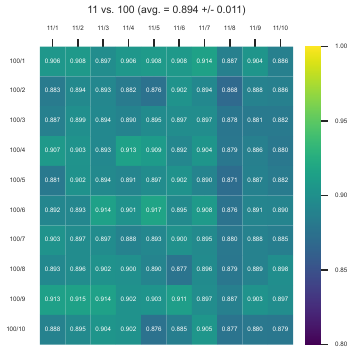}
         \caption{}
     \end{subfigure}
     \hfill
     \begin{subfigure}[b]{0.33\textwidth}
         \centering
         \includegraphics[width=\textwidth]{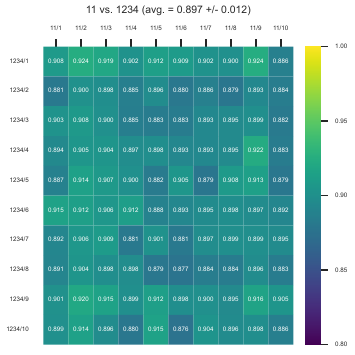}
         \caption{}
     \end{subfigure}
     \hspace{0.17\textwidth}
     \begin{subfigure}[b]{0.34\textwidth}
         \centering
         \includegraphics[width=\textwidth]{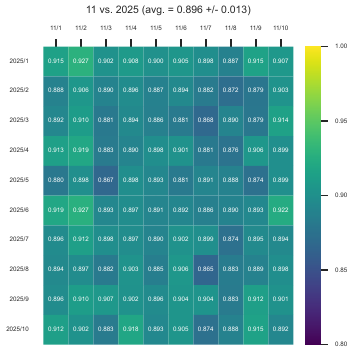}
         \caption{}
     \end{subfigure}
     \caption{Cosine similarities between the $10$ runs of module extraction of the chosen locality placement and all other locality placement runs, MNIST UMAP embeddings, $o=12$, $M=9$.}
     \label{fig:mnist_module_stability_across_loc_plac_per_loc}
\end{figure}

\begin{figure}[h]
     \centering
     \begin{subfigure}[b]{0.33\textwidth}
         \centering
         \includegraphics[width=\textwidth]{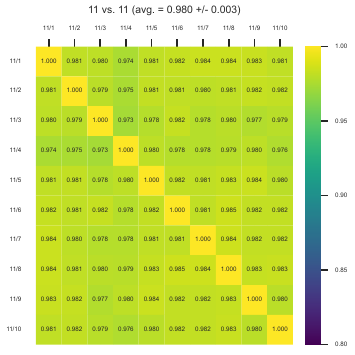}
         \caption{}
     \end{subfigure}
     \hfill
     \begin{subfigure}[b]{0.33\textwidth}
         \centering
         \includegraphics[width=\textwidth]{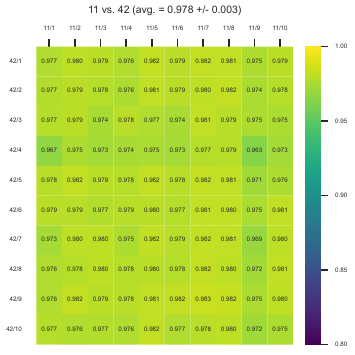}
         \caption{}
     \end{subfigure}
     \hfill
     \begin{subfigure}[b]{0.33\textwidth}
         \centering
         \includegraphics[width=\textwidth]{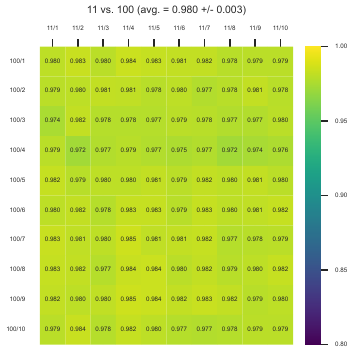}
         \caption{}
     \end{subfigure}
     \hfill
     \begin{subfigure}[b]{0.33\textwidth}
         \centering
         \includegraphics[width=\textwidth]{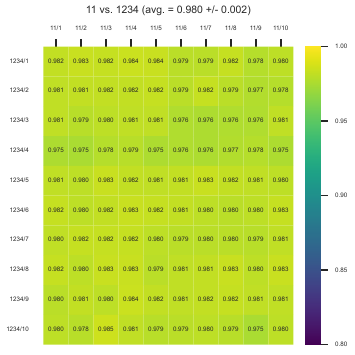}
         \caption{}
     \end{subfigure}
     \hspace{0.17\textwidth}
     \begin{subfigure}[b]{0.33\textwidth}
         \centering
         \includegraphics[width=\textwidth]{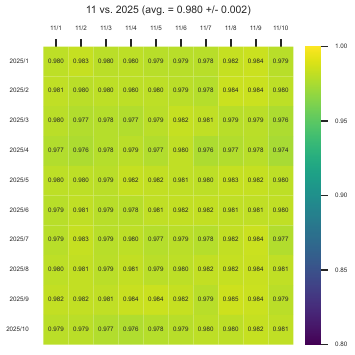}
         \caption{}
     \end{subfigure}
     \caption{Cosine similarities between the $10$ runs of module extraction of the chosen locality placement and all other locality placement runs, KPMP UMAP embeddings, $o=3$, $M=9$.}
     \label{fig:kpmp_module_stability_across_loc_plac_per_loc}
\end{figure}

To expand on results from Figure \ref{fig:stability}c, instead of averaging the correlation patterns the modules extract across different locality placements like we did there, here we show the comparisons of the presence-weighted module averages between all of the module runs, separately, from the chosen locality placement (denoted by its random seed $11$) and the other $4$ locality placement runs (Figure \ref{fig:mnist_module_stability_across_loc_plac_per_loc} for MNIST, and Figure \ref{fig:kpmp_module_stability_across_loc_plac_per_loc} for KPMP). The high cosine similarities indicate that the correlation patterns that module extraction explains are very similar across the different locality placements, even between individual module extraction runs, further highlighting the stability of our solutions. We also note that the cosine similarities between the presence-weighted module averages within one locality placement run are not significantly different than across the locality placement runs (e.g., compare figures \ref{fig:mnist_module_stability_across_loc_plac_per_loc}a and \ref{fig:mnist_module_stability_across_loc_plac_per_loc}c).

\begin{figure}[h]
     \centering
     \begin{subfigure}[b]{0.33\textwidth}
         \centering
         \includegraphics[width=\textwidth]{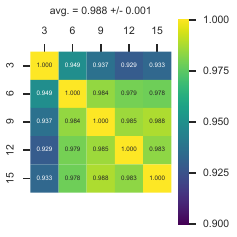}
         \caption{MNIST UMAP embeddings.}
     \end{subfigure}
     \hspace{0.17\textwidth}
     \begin{subfigure}[b]{0.33\textwidth}
         \centering
         \includegraphics[width=\textwidth]{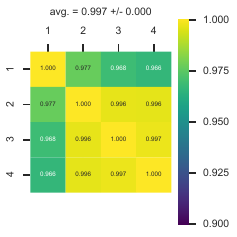}
         \caption{KPMP UMAP embeddings.}
     \end{subfigure}
     \caption{Cosine similarities between the average presence-weighted module averages across different overlap parameters $o$.}
     \label{fig:module_stab_across_overrep_o}
\end{figure}

Furthermore, we also look at the end-to-end stability of the module extraction when we change the overlap parameter $o$ as we did in Figure \ref{fig:stability}a. For each of the parameters $o$, we take averages of the presence-weighted module average across the $10$ runs for a single locality placement, and then compare between the different values of $o$ (Figure \ref{fig:module_stab_across_overrep_o}). These results further expand results demonstrate that the similarities of the captured correlations are not high only within different locality placements for some hyperparameter $o$, but also across different hyperparameters, and that these correlations are preserved and explained by our module extraction.

\clearpage

\subsection{MNIST Experiments}
\label{app:mnist_exp}

Here, we show some additional experiments and results done using the MNIST dataset.

\subsubsection{UMAP Embeddings: Local Patterns, Global Subpatterns}
\begin{figure}[h]
    \centering
    \includegraphics[width=\linewidth]{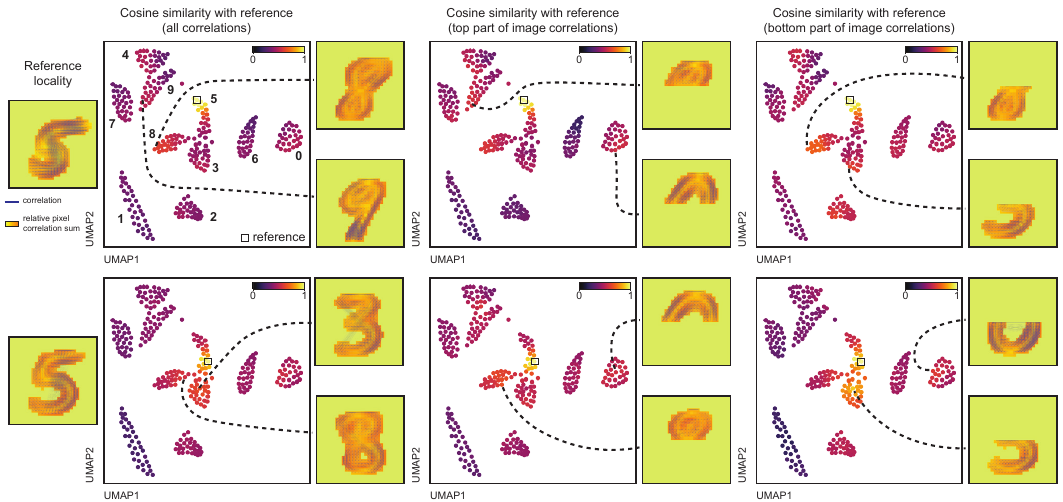}
    \caption{Cosine similarities between each of two reference localities and all other localities of the latent MNIST UMAP space in three settings, with example visualizations of similar localities.}
    \label{fig:locality_pattern_similarity}
\end{figure}

With this experiment, we explore the similarity between correlation patterns across a latent space when looking at different subsets of variables, to highlight the utility of extracting modules.
\textbf{Setup.} We use LAVA locality representations of the MNIST UMAP, without extracted modules. As examples, we select two different localities, and then calculate cosine similarity between their and other localities' respective representations. This is done using correlations for all feature pairs, or using subsets involving only features from the top or bottom half of the images.
\textbf{Intra- and intercluster heterogeneity.} We observe that patterns of correlations between the original features can be highly dissimilar across localities, even within the same cluster of digits (first column of Figure~\ref{fig:locality_pattern_similarity}). This can be conceptualized as a multivariate version of Simpson's paradox, wherein different groupings of data -- here LAVA localities -- can show different trends in association~\cite{samuels_simpsons_1993}, and highlights the fact that local feature associations cannot be trivially summarized across an entire dataset. \textbf{Subpatterns of correlations alter similarity.} 
Considering only locality correlations between pixels from the top or bottom half of the images, we observe variations in the cosine similarities across localities (second and third columns of Figure~\ref{fig:locality_pattern_similarity}). This indicates how different localities across the latent space might share specific subpatterns of correlation, but not others, which is precisely what the module extraction in LAVA aims to capture.

\subsubsection{Additional Results for UMAP embeddings}
\begin{figure}[h]
    \centering
    \includegraphics[width=0.5\linewidth]{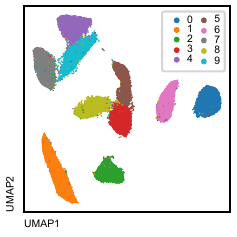}
    \caption{MNIST digit labels across UMAP.}
    \label{fig:mnist_meta_data}
\end{figure}

\begin{figure}[h]
     \centering
     \begin{subfigure}[b]{0.49\textwidth}
         \centering
         \includegraphics[width=\textwidth]{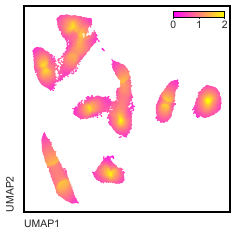}
         \caption{Scaled sample neighborhood in-degree centrality.}
     \end{subfigure}
     \hfill
     \begin{subfigure}[b]{0.49\textwidth}
         \centering
         \includegraphics[width=\textwidth]{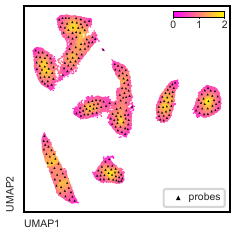}
         \caption{Scaled locality in-degree centrality, with probes.}
     \end{subfigure}
     \caption{MNIST UMAP sample relative centralities before and after locality placement (scaled by $\frac{E}{k}$ to have an average of $1$).}
     \label{fig:mnist_probes}
\end{figure}

\begin{figure}[h]
     \centering
     \begin{subfigure}[b]{0.32\textwidth}
         \centering
         \includegraphics[width=\textwidth]{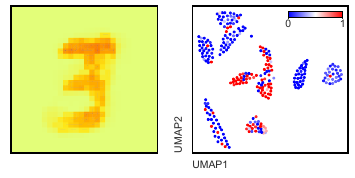}
         \caption{Module 1.}
     \end{subfigure}
     \hfill
     \begin{subfigure}[b]{0.32\textwidth}
         \centering
         \includegraphics[width=\textwidth]{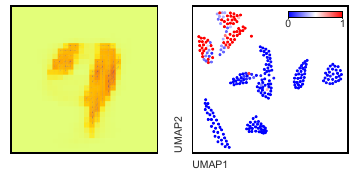}
         \caption{Module 2.}
     \end{subfigure}
     \hfill
     \begin{subfigure}[b]{0.32\textwidth}
         \centering
         \includegraphics[width=\textwidth]{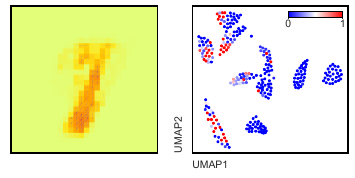}
         \caption{Module 3.}
     \end{subfigure}
     \hfill
     \begin{subfigure}[b]{0.32\textwidth}
         \centering
         \includegraphics[width=\textwidth]{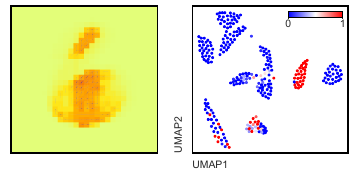}
         \caption{Module 4.}
     \end{subfigure}
     \hfill
     \begin{subfigure}[b]{0.32\textwidth}
         \centering
         \includegraphics[width=\textwidth]{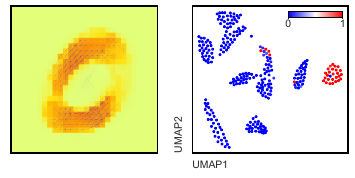}
         \caption{Module 5.}
     \end{subfigure}
     \hfill
     \begin{subfigure}[b]{0.32\textwidth}
         \centering
         \includegraphics[width=\textwidth]{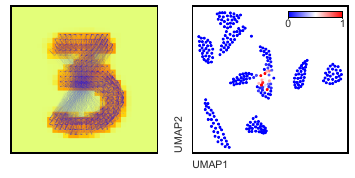}
         \caption{Module 6.}
     \end{subfigure}
     \hfill
     \begin{subfigure}[b]{0.32\textwidth}
         \centering
         \includegraphics[width=\textwidth]{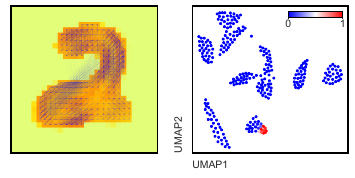}
         \caption{Module 7.}
     \end{subfigure}
     \hfill
     \begin{subfigure}[b]{0.32\textwidth}
         \centering
         \includegraphics[width=\textwidth]{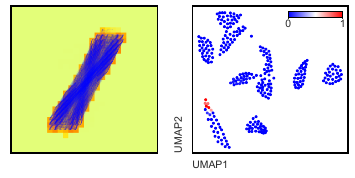}
         \caption{Module 8.}
     \end{subfigure}
     \hfill
     \begin{subfigure}[b]{0.32\textwidth}
         \centering
         \includegraphics[width=\textwidth]{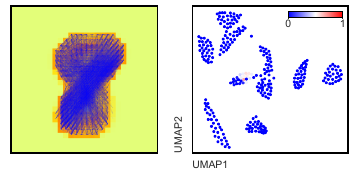}
         \caption{Module 9.}
     \end{subfigure}
     \caption{Visualizations and presences of the 9 extracted modules for MNIST.}
     \label{fig:mnist_modules_and_presences}
\end{figure}

\begin{figure}[h]
     \centering
     \begin{subfigure}[b]{0.32\textwidth}
         \centering
         \includegraphics[width=\textwidth]{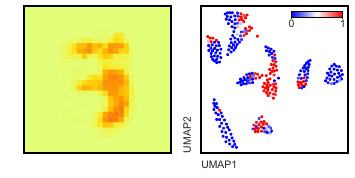}
         \caption{Module 1.}
     \end{subfigure}
     \hfill
     \begin{subfigure}[b]{0.32\textwidth}
         \centering
         \includegraphics[width=\textwidth]{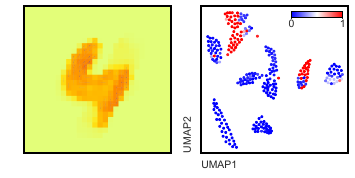}
         \caption{Module 2.}
     \end{subfigure}
     \hfill
     \begin{subfigure}[b]{0.32\textwidth}
         \centering
         \includegraphics[width=\textwidth]{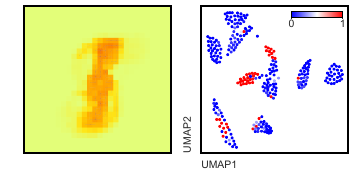}
         \caption{Module 3.}
     \end{subfigure}
     \hfill
     \begin{subfigure}[b]{0.32\textwidth}
         \centering
         \includegraphics[width=\textwidth]{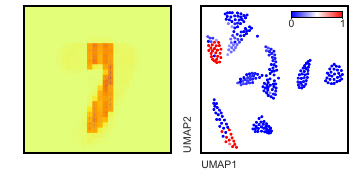}
         \caption{Module 4.}
     \end{subfigure}
     \hfill
     \begin{subfigure}[b]{0.32\textwidth}
         \centering
         \includegraphics[width=\textwidth]{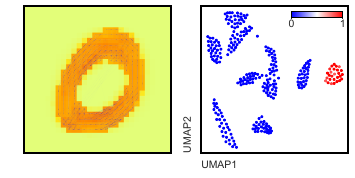}
         \caption{Module 5.}
     \end{subfigure}
     \hfill
     \begin{subfigure}[b]{0.32\textwidth}
         \centering
         \includegraphics[width=\textwidth]{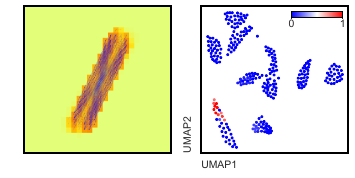}
         \caption{Module 6.}
     \end{subfigure}
     \hfill
     \begin{subfigure}[b]{0.32\textwidth}
         \centering
         \includegraphics[width=\textwidth]{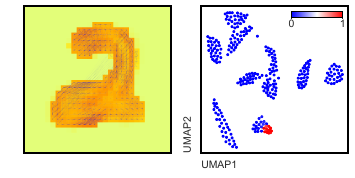}
         \caption{Module 7.}
     \end{subfigure}
     \hfill
     \begin{subfigure}[b]{0.32\textwidth}
         \centering
         \includegraphics[width=\textwidth]{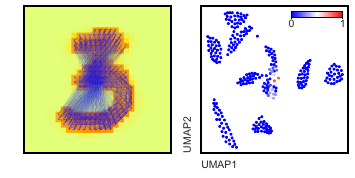}
         \caption{Module 8.}
     \end{subfigure}
     \hfill
     \begin{subfigure}[b]{0.32\textwidth}
         \centering
         \includegraphics[width=\textwidth]{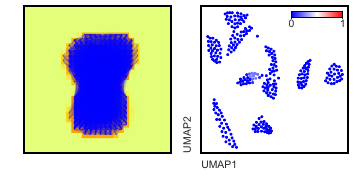}
         \caption{Module 9.}
     \end{subfigure}
     \caption{Visualizations and presences of the second best 9 modules for MNIST.}
     \label{fig:mnist_modules_and_presences_second}
\end{figure}

\begin{figure}[h]
     \centering
     \begin{subfigure}[b]{0.32\textwidth}
         \centering
         \includegraphics[width=\textwidth]{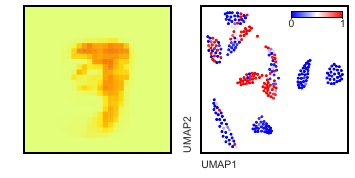}
         \caption{Module 1.}
     \end{subfigure}
     \hfill
     \begin{subfigure}[b]{0.32\textwidth}
         \centering
         \includegraphics[width=\textwidth]{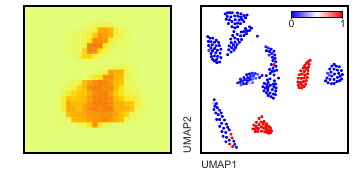}
         \caption{Module 2.}
     \end{subfigure}
     \hfill
     \begin{subfigure}[b]{0.32\textwidth}
         \centering
         \includegraphics[width=\textwidth]{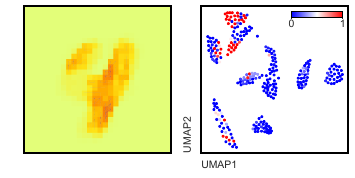}
         \caption{Module 3.}
     \end{subfigure}
     \hfill
     \begin{subfigure}[b]{0.32\textwidth}
         \centering
         \includegraphics[width=\textwidth]{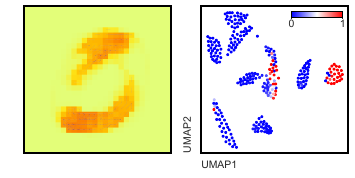}
         \caption{Module 4.}
     \end{subfigure}
     \hfill
     \begin{subfigure}[b]{0.32\textwidth}
         \centering
         \includegraphics[width=\textwidth]{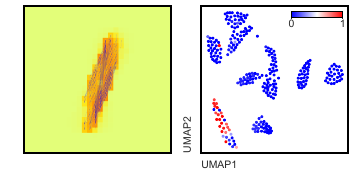}
         \caption{Module 5.}
     \end{subfigure}
     \hfill
     \begin{subfigure}[b]{0.32\textwidth}
         \centering
         \includegraphics[width=\textwidth]{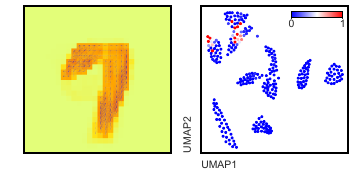}
         \caption{Module 6.}
     \end{subfigure}
     \hfill
     \begin{subfigure}[b]{0.32\textwidth}
         \centering
         \includegraphics[width=\textwidth]{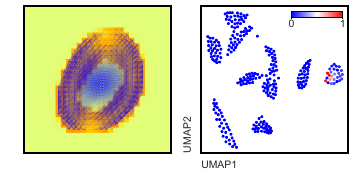}
         \caption{Module 7.}
     \end{subfigure}
     \hfill
     \begin{subfigure}[b]{0.32\textwidth}
         \centering
         \includegraphics[width=\textwidth]{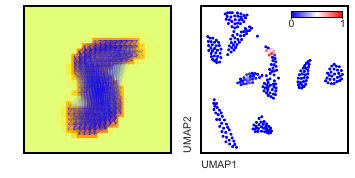}
         \caption{Module 8.}
     \end{subfigure}
     \hfill
     \begin{subfigure}[b]{0.32\textwidth}
         \centering
         \includegraphics[width=\textwidth]{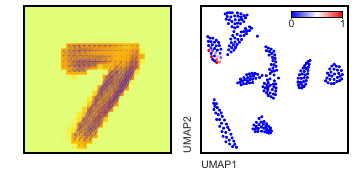}
         \caption{Module 9.}
     \end{subfigure}
     \caption{Visualizations and presences of the third best 9 modules for MNIST.}
     \label{fig:mnist_modules_and_presences_third}
\end{figure}

In Figure \ref{fig:mnist_meta_data}, we show the class labels of individual samples in our UMAP embeddings. Then, in Figure \ref{fig:mnist_probes}, we show the relative in-degree centralities of samples when looking at all the neighborhoods and when looking at our locality placement. In Figure \ref{fig:mnist_modules_and_presences}, we visualize all modules and their presences used in our experiments from Section \ref{sec:experiments}. We also show modules and presences for the second and third best solutions when $L=9$ (Figure \ref{fig:mnist_modules_and_presences_second} and Figure \ref{fig:mnist_modules_and_presences_third}).

\begin{figure}[h]
     \centering
     \begin{subfigure}[b]{0.32\textwidth}
         \centering
         \includegraphics[width=\textwidth]{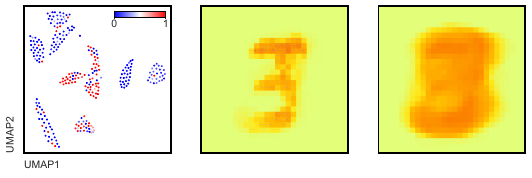}
         \caption{Module 1.}
     \end{subfigure}
     \hfill
     \begin{subfigure}[b]{0.32\textwidth}
         \centering
         \includegraphics[width=\textwidth]{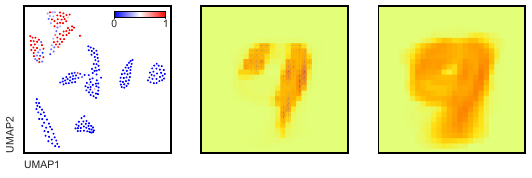}
         \caption{Module 2.}
     \end{subfigure}
     \hfill
     \begin{subfigure}[b]{0.32\textwidth}
         \centering
         \includegraphics[width=\textwidth]{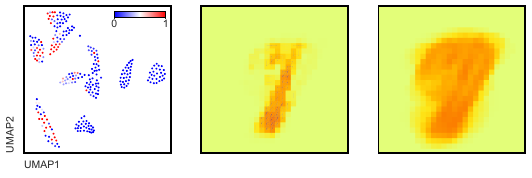}
         \caption{Module 3.}
     \end{subfigure}
     \hfill
     \begin{subfigure}[b]{0.32\textwidth}
         \centering
         \includegraphics[width=\textwidth]{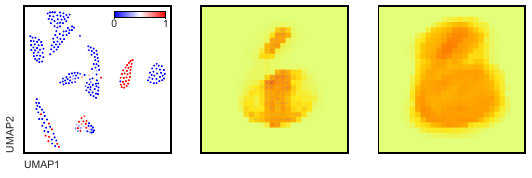}
         \caption{Module 4.}
     \end{subfigure}
     \hfill
     \begin{subfigure}[b]{0.32\textwidth}
         \centering
         \includegraphics[width=\textwidth]{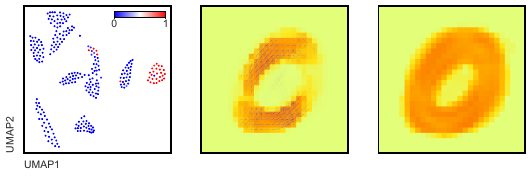}
         \caption{Module 5.}
     \end{subfigure}
     \hfill
     \begin{subfigure}[b]{0.32\textwidth}
         \centering
         \includegraphics[width=\textwidth]{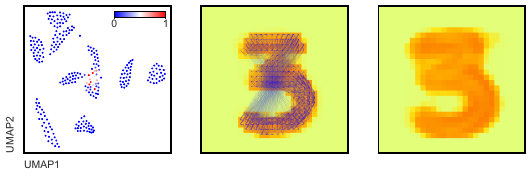}
         \caption{Module 6.}
     \end{subfigure}
     \hfill
     \begin{subfigure}[b]{0.32\textwidth}
         \centering
         \includegraphics[width=\textwidth]{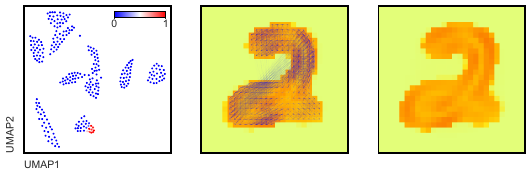}
         \caption{Module 7.}
     \end{subfigure}
     \hfill
     \begin{subfigure}[b]{0.32\textwidth}
         \centering
         \includegraphics[width=\textwidth]{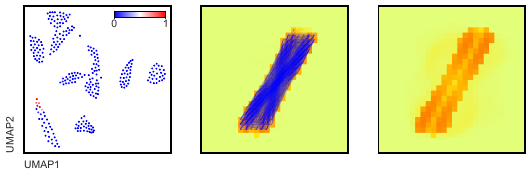}
         \caption{Module 8.}
     \end{subfigure}
     \hfill
     \begin{subfigure}[b]{0.32\textwidth}
         \centering
         \includegraphics[width=\textwidth]{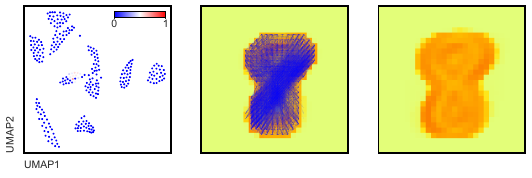}
         \caption{Module 9.}
     \end{subfigure}
     \caption{Presences, actual extracted modules, and what modules would look like if they were presence-weighted averages of the localities they reconstruct, for the $9$ extracted MNIST modules.}
     \label{fig:mnist_modules_all_simulated_averages}
\end{figure}

Finally, in Figure \ref{fig:mnist_modules_all_simulated_averages}, we also visualize how the chosen modules would look like if they were the presence-weighted average of their localities' correlations, rather than overlapping patterns we extract with AMF (see Section \ref{sec:experiments}). While both the presences and modules would change if we were to alter our loss function (they are optimized jointly), the figure still illustrates how the overestimation weighing helps make sure modules we extract are sparse correlation subpatterns contained in all of the localities the modules are present in, rather than an average correlation across those localities.

\clearpage

\subsubsection{Additional Results for Autoencoder Embeddings}
\label{app:additional_autoencoder}
Here we show some extra results based on the autoencoder embeddings. In visualizations for these embeddings of MNIST, we use a 2-dimensional UMAP (default hyperparameters) of the original 4-dimensional latent space. The UMAP is generated using only the embeddings, and to show a probe location, we use the sample nearest to the probe in the original latent space. Note that all steps of LAVA were run on the original space, not the reduced 2-dimensional UMAP.

\begin{figure}[h]
    \centering
    \includegraphics[width=0.49\linewidth]{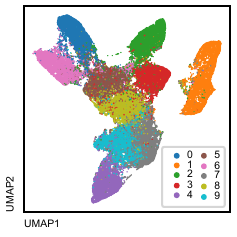}
    \caption{MNIST digit labels across the autoencoder embeddings (UMAP visualization).}
    \label{fig:mnist_meta_data_ae}
\end{figure}

\begin{figure}[h]
     \centering
     \begin{subfigure}[b]{0.49\textwidth}
         \centering
         \includegraphics[width=\textwidth]{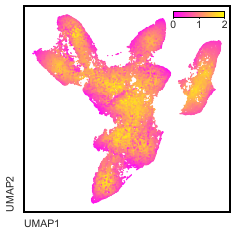}
         \caption{Scaled sample neighborhood in-degree centrality.}
     \end{subfigure}
     \hfill
     \begin{subfigure}[b]{0.49\textwidth}
         \centering
         \includegraphics[width=\textwidth]{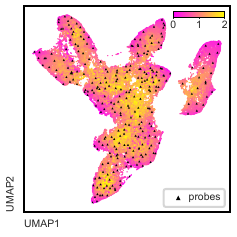}
         \caption{Scaled locality in-degree centrality, with probes.}
     \end{subfigure}
     \caption{MNIST autoencoder sample relative centralities before and after locality placement (scaled by $\frac{E}{k}$ to have an average of $1$; visualized using a UMAP of the original embeddings).}
     \label{fig:mnist_probes_ae}
\end{figure}

\begin{figure}[h]
    \centering
    \includegraphics[width=0.5\linewidth]{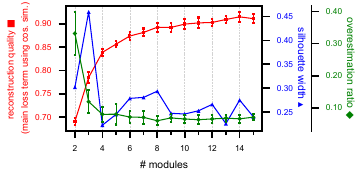}
    \caption{Module extraction statistics for the MNIST autoencoder embeddings.}
    \label{fig:mnist_stability_ae}
\end{figure}

\begin{figure}[h]
     \centering
     \begin{subfigure}[b]{0.32\textwidth}
         \centering
         \includegraphics[width=\textwidth]{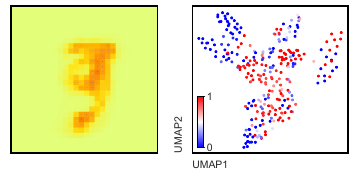}
         \caption{Module 1.}
     \end{subfigure}
     \hfill
     \begin{subfigure}[b]{0.32\textwidth}
         \centering
         \includegraphics[width=\textwidth]{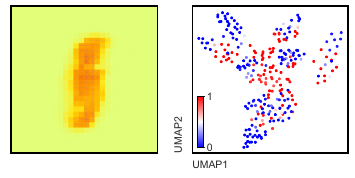}
         \caption{Module 2.}
     \end{subfigure}
     \hfill
     \begin{subfigure}[b]{0.32\textwidth}
         \centering
         \includegraphics[width=\textwidth]{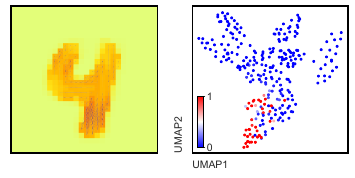}
         \caption{Module 3.}
     \end{subfigure}
     \hfill
     \begin{subfigure}[b]{0.32\textwidth}
         \centering
         \includegraphics[width=\textwidth]{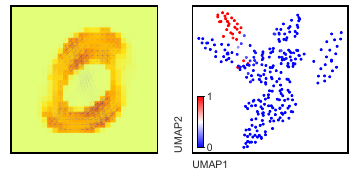}
         \caption{Module 4.}
     \end{subfigure}
     \hfill
     \begin{subfigure}[b]{0.32\textwidth}
         \centering
         \includegraphics[width=\textwidth]{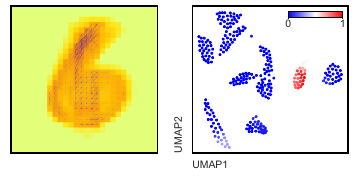}
         \caption{Module 5.}
     \end{subfigure}
     \hfill
     \begin{subfigure}[b]{0.32\textwidth}
         \centering
         \includegraphics[width=\textwidth]{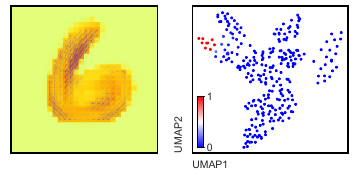}
         \caption{Module 6.}
     \end{subfigure}
     \hfill
     \begin{subfigure}[b]{0.32\textwidth}
         \centering
         \includegraphics[width=\textwidth]{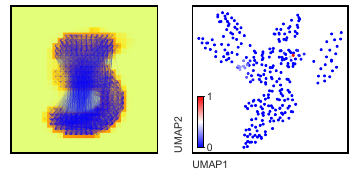}
         \caption{Module 7.}
     \end{subfigure}
     \begin{subfigure}[b]{0.32\textwidth}
         \centering
         \includegraphics[width=\textwidth]{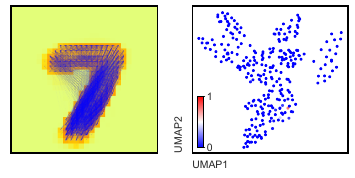}
         \caption{Module 8.}
     \end{subfigure}
     \caption{Visualizations and presences of the chosen 8 modules for MNIST autoencoder embeddings.}
     \label{fig:mnist_modules_and_presences_ae}
\end{figure}

In Figure \ref{fig:mnist_meta_data_ae}, we show the class label distribution of the latent space. Then, in Figure \ref{fig:mnist_probes_ae}, we show the relative in-degree centralities of samples when looking at all the neighborhoods and when looking at our locality placement. In Figure \ref{fig:mnist_stability_ae} we show the results of running module extraction using a different number modules. Based on these results, we selected the runs with $8$ modules, and then used the module extraction run with the lowest main loss term, like in other experiments. In Figure \ref{fig:mnist_modules_and_presences_ae}, we visualize all the $8$ modules and their presences from the best run, which were used in our LFXAI comparison experiments in Section \ref{sec:experiments}.

\clearpage

\subsubsection{Additional Results for Comparison with LFXAI}
\label{app:additional_lfxai} 

\begin{figure}[h]
     \centering
     \begin{subfigure}[b]{0.32\textwidth}
         \centering
         \includegraphics[width=\textwidth]{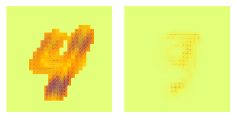}
         \caption{Locality correlations and LAVA reconstruction from example locality 1.}
     \end{subfigure}
     \hfill
     \begin{subfigure}[b]{0.32\textwidth}
         \centering
         \includegraphics[width=\textwidth]{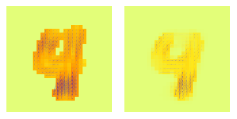}
         \caption{Locality correlations and LAVA reconstruction from example locality 2.}
     \end{subfigure}
     \hfill
     \begin{subfigure}[b]{0.32\textwidth}
         \centering
         \includegraphics[width=\textwidth]{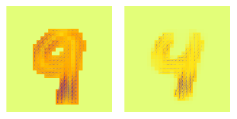}
         \caption{Locality correlations and LAVA reconstruction from example locality 3.}
     \end{subfigure}
     \caption{Visualizations of localities used as examples in Figure \ref{sec:lfxai_comp}, alongside their reconstructions using LAVA modules.}
     \label{fig:lfxai_lava_example_locality_and_recon}
\end{figure}

\begin{figure}[h]
     \centering
     \begin{subfigure}[b]{0.32\textwidth}
         \centering
         \includegraphics[width=\textwidth]{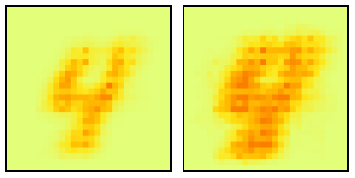}
         \caption{Locality-averaged gradSHAP feature importance and its standard deviation for example locality 1.}
     \end{subfigure}
     \hfill
     \begin{subfigure}[b]{0.32\textwidth}
         \centering
         \includegraphics[width=\textwidth]{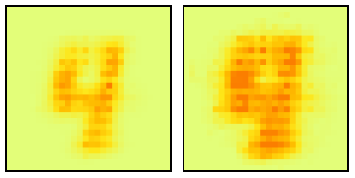}
         \caption{Locality-averaged gradSHAP feature importance and its standard deviation for example locality 2.}
     \end{subfigure}
     \hfill
     \begin{subfigure}[b]{0.32\textwidth}
         \centering
         \includegraphics[width=\textwidth]{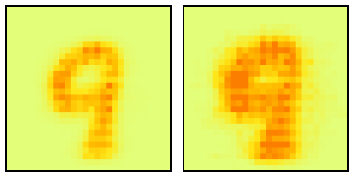}
         \caption{Locality-averaged gradSHAP feature importance and its standard deviation for example locality 3.}
     \end{subfigure}
     \caption{Visualizations of the locality-averaged LFXAI-based gradSHAP feature importance explanations, alongside their standard deviations.}
     \label{fig:lfxai_average_with_std}
\end{figure}

\begin{figure}[h]
     \centering
     \begin{subfigure}[b]{0.26\textwidth}
         \centering
         \includegraphics[width=\textwidth]{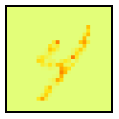}
         \caption{Feature importance of a sample from example locality 1 (1).}
     \end{subfigure}
     \hfill
     \begin{subfigure}[b]{0.26\textwidth}
         \centering
         \includegraphics[width=\textwidth]{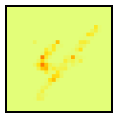}
         \caption{Feature importance of a sample from example locality 1 (2).}
     \end{subfigure}
     \hfill
     \begin{subfigure}[b]{0.26\textwidth}
         \centering
         \includegraphics[width=\textwidth]{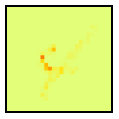}
         \caption{Feature importance of a sample from example locality 1 (3).}
     \end{subfigure}
     \hfill
     \begin{subfigure}[b]{0.26\textwidth}
         \centering
         \includegraphics[width=\textwidth]{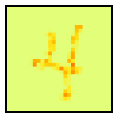}
         \caption{Feature importance of a sample from example locality 2 (1).}
     \end{subfigure}
     \hfill
     \begin{subfigure}[b]{0.26\textwidth}
         \centering
         \includegraphics[width=\textwidth]{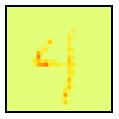}
        \caption{Feature importance of a sample from example locality 2 (2).}
     \end{subfigure}
     \hfill
     \begin{subfigure}[b]{0.26\textwidth}
         \centering
         \includegraphics[width=\textwidth]{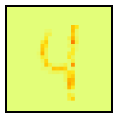}
         \caption{Feature importance of a sample from example locality 2 (3).}
     \end{subfigure}
     \hfill
     \begin{subfigure}[b]{0.26\textwidth}
         \centering
         \includegraphics[width=\textwidth]{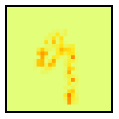}
         \caption{Feature importance of a sample from example locality 3 (1).}
     \end{subfigure}
    \hfill
     \begin{subfigure}[b]{0.26\textwidth}
         \centering
         \includegraphics[width=\textwidth]{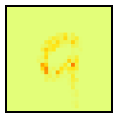}
         \caption{Feature importance of a sample from example locality 3 (2).}
     \end{subfigure}
     \hfill
     \begin{subfigure}[b]{0.26\textwidth}
         \centering
         \includegraphics[width=\textwidth]{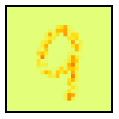}
         \caption{Feature importance of a sample from example locality 3 (3).}
     \end{subfigure}
     \caption{Visualizations of LFXAI-based gradSHAP feature importance for the three example localities used in Figure \ref{fig:lfxai_comparison}.}
     \label{fig:mnist_lfxai_single_sample_examples}
\end{figure}

Here we expand on Section \ref{sec:lfxai_comp}, and especially Figure \ref{fig:lfxai_comparison}. In Figure \ref{fig:lfxai_lava_example_locality_and_recon} we show visualize the correlation patterns of the three example localities alongside their module-based reconstructions. In Figure \ref{fig:lfxai_average_with_std} we show average LFXAI-based feature importance, alongside their standard deviation (on the same scale as the importance): we can see that standard deviations are generally larger than the importance themselves for these embeddings, making it unclear whether any consistently relevant patterns can be observed in this case. This could be due to the shift equivariant nature of a convolutional autoencoder. We also show a couple of examples of the feature importance before averaging to visualize where the variance comes from (Figure \ref{fig:mnist_lfxai_single_sample_examples}). We additionally note that feature importance used an empty image as a reference (all features were $0$, i.e., pixels were white). Another reference could have been used, e.g., random subset of the dataset, or some other reference, but we did not see how altering the reference used could make these explanations more comparable to LAVA without significant alterations.

\clearpage

\subsection{Comparing AMF to NMF}
\label{app:nmf_amf}
Here we compare AMF to Non-negative Matrix Factorization (NMF), the method which we found to be conceptually most similar, and therefore most relevant to motivate the need for developing a new method~\cite{lee_learning_1999}.

The aim of step 3 in LAVA is to capture subpatterns of correlation across latent embeddings, as defined through localities. While in principle it might appear suitable to use a more general factorization approach, AMF was designed with the following additional properties in mind: to capture reoccurring rather than averaged subpatterns of correlation across latent embeddings, to focus on patterns rather than absolute values, and to focus on localities with many high correlations as opposed to few or mostly weak correlations. These aims led to the corresponding modeling choices for AMF, motivated in Section \ref{sec:methods}, which also make it different from other matrix factorization methods.

\subsubsection{Conceptual Differences between AMF and NMF}
NMF decomposes some non-negative matrix $V$ $(n \times m)$ into non-negative matrices $W$ $(n \times k)$ and $H$ $(k \times m)$, where $k$ is a hyperparameter that determines the number of “components” or “parts”, and $W$ and $H$ are optimized so that $V = WH$\footnote{Using standard NMF naming convention for the different variables, not to be confused with any of the variable names from our method.}. While NMF and AMF are both conceptually “parts-based” matrix decompositions, we highlight three distinct features of AMF: (1) entries of the AMF module and presence matrices are restricted to real values in the $[0, 1]$, whereas in NMF matrices are restricted only to non-negative values; (2) the AMF method uses outer products and a maximum function to reconstruct the matrix (localities with their corresponding correlations), rather than standard matrix multiplication like in NMF (Section \ref{sec:module_extraction}; Figures \ref{fig:module_extraction} and \ref{fig:overlap}); (3) AMF uses a similarity-based loss function that incorporates a quantile-regression-like weighing scheme, rather than optimizing some form of beta-divergence as is the case with NMF. We motivate these essential modeling choices throughout Section \ref{sec:module_extraction} in terms of the desired properties of the modules, and we showcase the effectiveness of our formulation in terms of module (and presence) properties in Section \ref{sec:module_exp}.

\subsubsection{Empirical Results of Using NMF to Extract modules}
To illustrate some of the differences of the parts/modules extracted by the two methods, we apply NMF to the LAVA localities of the UMAP embeddings of MNIST and KPMP (from our main paper experiments). For NMF, we used random initialization and the KL divergence as the loss for NMF. We deemed KL divergence as the most similar beta-divergence in relation to our cosine distance loss function.

\begin{figure}[h]
     \centering
     \begin{subfigure}[b]{0.45\textwidth}
         \centering
         \includegraphics[width=\textwidth]{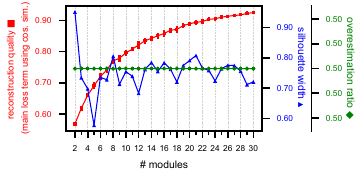}
         \caption{MNIST UMAP NMF module extraction.}
     \end{subfigure}
     \hfill
     \begin{subfigure}[b]{0.45\textwidth}
         \centering
         \includegraphics[width=\textwidth]{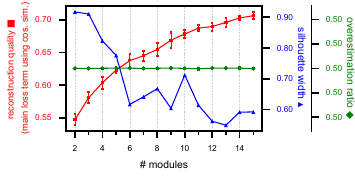}
         \caption{KPMP UMAP NMF module extraction.}
     \end{subfigure}
     \caption{Module extraction runs (10 runs per $\# \:modules$) using NMF for the UMAP embeddings of the MNIST and KPMP datasets.}
     \label{fig:nmf_mod_ex}
\end{figure}

\begin{figure}[h]
     \centering
     \begin{subfigure}[b]{0.49\textwidth}
         \centering
         \includegraphics[width=\textwidth]{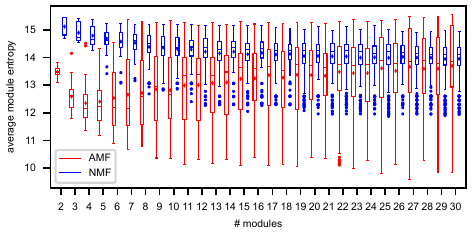}
         \caption{MNIST UMAP module entropies.}
     \end{subfigure}
     \begin{subfigure}[b]{0.49\textwidth}
         \centering
         \includegraphics[width=\textwidth]{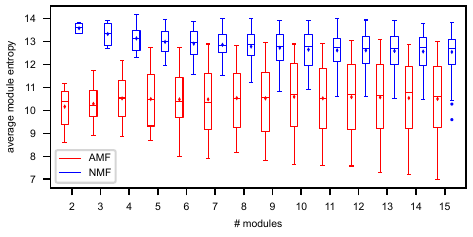}
         \caption{KPMP UMAP module entropies.}
     \end{subfigure}
     \caption{Shannon entropy of modules across the module extraction runs (10 module runs per $\# \:modules$).}
     \label{fig:nmf_amf_entr}
\end{figure}

\begin{figure}[h]
     \centering
     \begin{subfigure}[b]{0.32\textwidth}
         \centering
         \includegraphics[width=\textwidth]{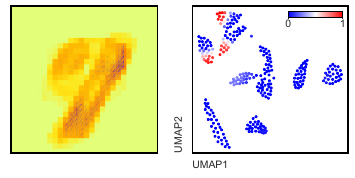}
         \caption{Module 1.}
     \end{subfigure}
     \hfill
     \begin{subfigure}[b]{0.32\textwidth}
         \centering
         \includegraphics[width=\textwidth]{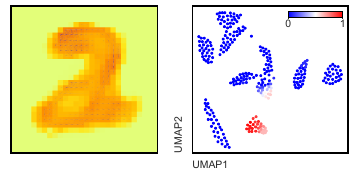}
         \caption{Module 2.}
     \end{subfigure}
     \hfill
     \begin{subfigure}[b]{0.32\textwidth}
         \centering
         \includegraphics[width=\textwidth]{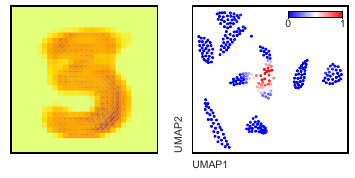}
         \caption{Module 3.}
     \end{subfigure}
     \hfill
     \begin{subfigure}[b]{0.32\textwidth}
         \centering
         \includegraphics[width=\textwidth]{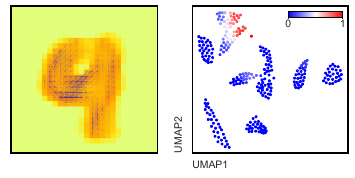}
         \caption{Module 4.}
     \end{subfigure}
     \hfill
     \begin{subfigure}[b]{0.32\textwidth}
         \centering
         \includegraphics[width=\textwidth]{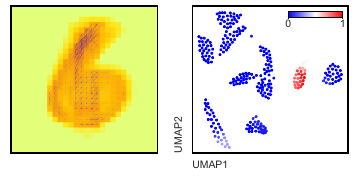}
         \caption{Module 5.}
     \end{subfigure}
     \hfill
     \begin{subfigure}[b]{0.32\textwidth}
         \centering
         \includegraphics[width=\textwidth]{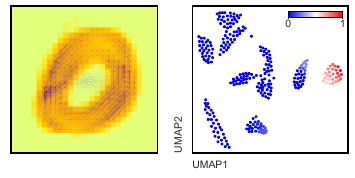}
         \caption{Module 6.}
     \end{subfigure}
     \hfill
     \begin{subfigure}[b]{0.32\textwidth}
         \centering
         \includegraphics[width=\textwidth]{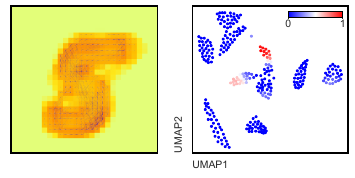}
         \caption{Module 7.}
     \end{subfigure}
     \hfill
     \begin{subfigure}[b]{0.32\textwidth}
         \centering
         \includegraphics[width=\textwidth]{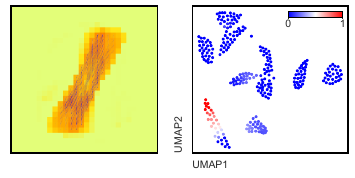}
         \caption{Module 8.}
     \end{subfigure}
     \hfill
     \begin{subfigure}[b]{0.32\textwidth}
         \centering
         \includegraphics[width=\textwidth]{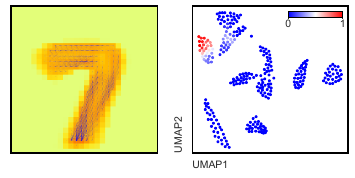}
         \caption{Module 9.}
     \end{subfigure}
     \caption{Visualizations of the NMF components and their coefficients across the LAVA-generated localities.}
     \label{fig:mnist_umap_nmf_modules_and_presences}
\end{figure}

Firstly, we report module extraction statistics when varying the number of modules/components in Figure \ref{fig:nmf_mod_ex} (we run NMF $10$ for each number of components, just like in Figure \ref{fig:stability}b). The first thing we note is that the overestimation ratio is consistently $0.5$, which means that if we look at NMF components that are used to reconstruct the various localities, on average half of the correlations in those parts will be overestimated and half underestimated, which makes it impossible to treat them as parts of the underlying correlations of localities, but rather as their averages (which is one of the things we prevent with our AMF formulation). We also notice that optimizing KL divergence is indeed similar to our cosine-distance-based loss, as the loss function inverse increases steadily with the increase of number of NMF parts (referred to as modules in the figure). However, NMF does yield higher silhouette widths, indicating that its formulation will produce more stable solutions. We already commented on why we think this is the case with AMF in Section \ref{sec:module_exp}. To further quantify the differences between the AMF modules and NMF components, we calculate the Shannon entropy of every extracted module for all of the module extraction runs above, and compare them between AMF and NMF (Figure \ref{fig:nmf_amf_entr}). A higher entropy indicates that the correlation pattern is more spread-out, while a lower entropy means that the pattern is sparser. We can see that NMF will consistently produce more spread-out correlations, coinciding with the fact that it does not optimize for lower overestimation errors and is therefore more likely to average across the localities rather than extract their subcomponents. Looking at the dynamics of the entropy values as we increase the number of modules, we see that AMF modules will start with sparser modules, and produce both more spread-out and sparser modules as we increase their number. Based on other results from our AMF-related experiments, we can infer that when using higher numbers of modules, the correlation patterns in our embeddings will produce some sparse but widely-reoccurring modules (low entropy modules), while the rest of the modules will be spent on capturing the correlations patterns of the more specific localities that do not contain these sparser patterns (high entropy modules). When looking at NMF in relation, we can see that it will produce less sparse components from the start, with some sparser components occurring more often only when we increase the number of components (still not as sparse as AMF). Finally, we also show some visualizations of a NMF-produced components. We use a run with $9$ modules whose reconstructions had the lowest AMF loss to enable a fair qualitative comparison with our AMF modules from Section \ref{sec:experiments} (Figure \ref{fig:mnist_umap_nmf_modules_and_presences}). The visualizations mainly confirm what we already observe quantitatively: while the more spread-out modules do manage to capture some wide-spread patterns of correlations, the fact that we do not control overestimation (e.g., module 5 that clearly averages across the correlation of localities containing images of the numbers $1$ and $6$) and the fact that NMF adds up components rather than overlaps them like AMF (Figure \ref{fig:overlap}c) makes interpretation of the NMF-captured correlation patterns virtually impossible.

We acknowledge the existence of other variants of NMFs, such as NMF with sparse constraints~\cite{o_non-negative_2004}, which might be more fitting to our task. However, we do not take it upon ourselves to perform additional experiments to evaluate these properties, maintaining that the conceptual differences we already highlighted are enough to demonstrate that any of these alternatives are still unfit for explaining variation of correlation patterns we find in the latent localities of LAVA. We leave for future work seeing what the relationship between sparsity and overestimation constraints is, and whether this could be used in alternative loss functions of the AMF method.

\clearpage

\subsection{KPMP Experiments}
\label{app:kpmp_exp}

Here, we show some extra figures and results relating to experiments done with the KPMP dataset, and we provide more details on the gene set analysis done in Section \ref{sec:kpmp_exp}.

\subsubsection{Additional figures and results.}

\begin{figure}[h]
     \centering
     \begin{subfigure}[b]{0.66\textwidth}
         \centering
         \includegraphics[width=\textwidth]{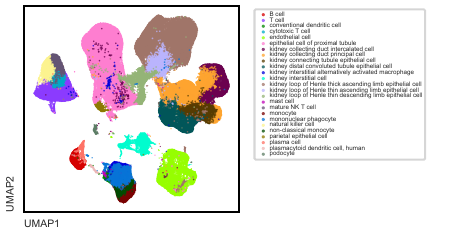}
         \caption{Cell types.}
     \end{subfigure}
     \hfill
     \begin{subfigure}[b]{0.33\textwidth}
         \centering
         \includegraphics[width=\textwidth]{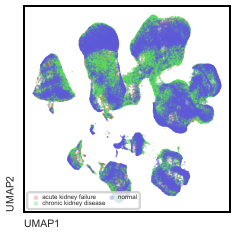}
         \caption{Disease states.}
     \end{subfigure}
     \caption{KPMP metadata.}
     \label{fig:kpmp_metadata}
\end{figure}

\begin{figure}[h]
     \centering
     \begin{subfigure}[b]{0.49\textwidth}
         \centering
         \includegraphics[width=\textwidth]{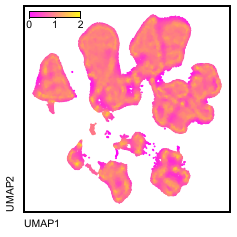}
         \caption{Scaled sample neighborhood in-degree centrality.}
     \end{subfigure}
     \hfill
     \begin{subfigure}[b]{0.49\textwidth}
         \centering
         \includegraphics[width=\textwidth]{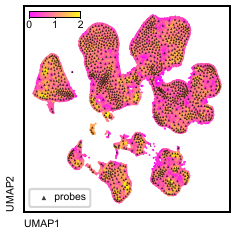}
         \caption{Scaled locality in-degree centrality, with probes.}
     \end{subfigure}
     \caption{KPMP UMAP relative sample centralities before and after locality placement (scaled by $\frac{E}{k}$ to have an average of $1$).}
     \label{fig:kpmp_probes}
\end{figure}

\begin{figure}[h]
     \centering
     \begin{subfigure}[b]{0.32\textwidth}
         \centering
         \includegraphics[width=\textwidth]{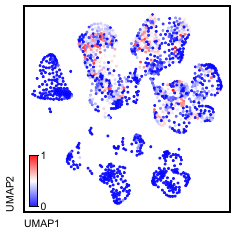}
         \caption{Module 1.}
     \end{subfigure}
     \hfill
     \begin{subfigure}[b]{0.32\textwidth}
         \centering
         \includegraphics[width=\textwidth]{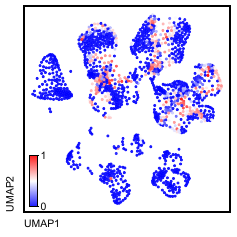}
         \caption{Module 2.}
     \end{subfigure}
     \hfill
     \begin{subfigure}[b]{0.32\textwidth}
         \centering
         \includegraphics[width=\textwidth]{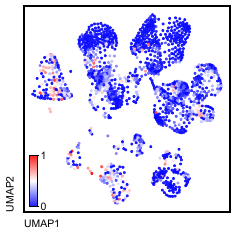}
         \caption{Module 3.}
     \end{subfigure}
     \hfill
     \begin{subfigure}[b]{0.32\textwidth}
         \centering
         \includegraphics[width=\textwidth]{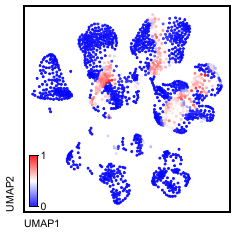}
         \caption{Module 4.}
     \end{subfigure}
     \hfill
     \begin{subfigure}[b]{0.32\textwidth}
         \centering
         \includegraphics[width=\textwidth]{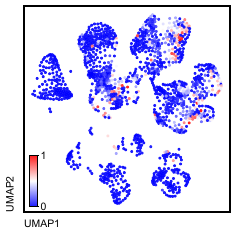}
         \caption{Module 5.}
     \end{subfigure}
     \hfill
     \begin{subfigure}[b]{0.32\textwidth}
         \centering
         \includegraphics[width=\textwidth]{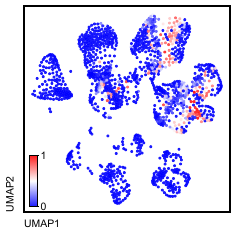}
         \caption{Module 6.}
     \end{subfigure}
     \hfill
     \begin{subfigure}[b]{0.32\textwidth}
         \centering
         \includegraphics[width=\textwidth]{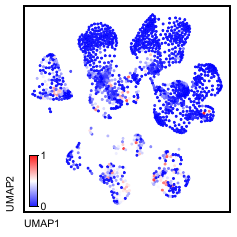}
         \caption{Module 7.}
     \end{subfigure}
     \hfill
     \begin{subfigure}[b]{0.32\textwidth}
         \centering
         \includegraphics[width=\textwidth]{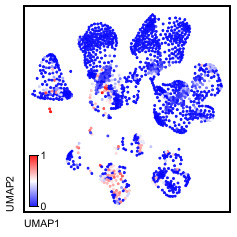}
         \caption{Module 8.}
     \end{subfigure}
     \hfill
     \begin{subfigure}[b]{0.32\textwidth}
         \centering
         \includegraphics[width=\textwidth]{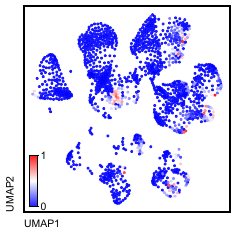}
         \caption{Module 9.}
     \end{subfigure}
     \caption{Presences of the 9 extracted modules for KPMP.}
     \label{fig:kpmp_presences}
\end{figure}

In Figure \ref{fig:kpmp_metadata}, we show cell type and disease category labels for all the samples in the UMAP embeddings. In Figure \ref{fig:kpmp_probes}, we show the relative in-degree centralities of samples when looking at all the neighborhoods and when looking at our locality placement. In Figure \ref{fig:kpmp_presences}, we show the presences of modules chosen for experiments (Section \ref{sec:kpmp_exp}).

\begin{figure}[h]
     \centering
     \begin{subfigure}[b]{0.3\textwidth}
         \centering
         \includegraphics[width=\textwidth]{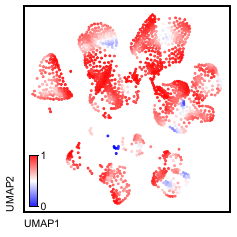}
         \caption{Any disease.}
     \end{subfigure}
     \hfill
     \begin{subfigure}[b]{0.3\textwidth}
         \centering
         \includegraphics[width=\textwidth]{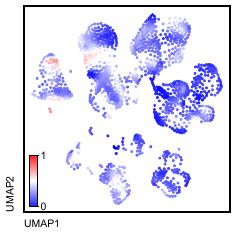}
         \caption{AKF: acute kidney failure.}
     \end{subfigure}
     \hfill
     \begin{subfigure}[b]{0.3\textwidth}
         \centering
         \includegraphics[width=\textwidth]{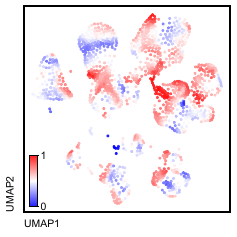}
         \caption{CKD: chronic kidney disease.}
     \end{subfigure}
        \caption{Percentage of disease categories across localities.}
        \label{fig:disease_status_percentages}
\end{figure}

\textbf{Disease category.} In Section \ref{sec:kpmp_exp}, we analyze our modules in relation to disease category of patients from whom the single-cell samples come from. To get a better sense of the data, in Figure \ref{fig:disease_status_percentages}, we visualize the ratios of diseased samples across LAVA's localities. \textbf{Correlations between module presences and disease category ratios.} For each of the modules, we also report correlations between the presences of that module and the ratios of the three different disease categories (Table \ref{tab:disease_correlations}). We calculate correlations only across localities in which the module is actually present, that is, with presence $> 0.01$. We do this because we do not expect that a pattern of correlations in a locality will be caused purely by the number of disease samples. Therefore, we also do not expect any module presences to directly reflect the number of disease samples across the localities. However, we do expect that the presence of a module might reflect the relative number of disease samples across the localities it is present in, as a disease could have a specific influence on gene expression for specific cells, and hence also the correlations we might observe when looking at the embeddings locally.

\begin{table}[h]
    \tiny
    \centering
    \begin{tabular}{|c||c|c|c|c|c|c|} \hline  
 & CKD + AKF, Pear.& CKD + AKF, Spear.& CKD, Pear.& CKD, Spear.& AKF, Pear.& AKF, Spear.\\ \hline \hline
         m. 1&  $-.06$ ($p=.116$)&  $-.06$ ($p=.139$)&  $-.11$ ($p=.005$)&  $-.08$ ($p=.035$)&  $.07$ ($p=.049$)&  $.07$ ($p=.052$)\\ \hline  
         m. 2&  $.01$ ($p=.865$)&  $.03$ ($p=.473$)&  $-.03$ ($p=.592$)&  $-.02.$ ($p=.62$)&  $.05$ ($p=.292$)&  $.08$ ($p=.095$)\\ \hline  
         m. 3&  $-.01$ ($p=.878$)&  $-.05$ ($p=.242$)&  $-.18$ ($p<.001$)&  $-.20$ ($p=$)&  $.23$ ($p<.001$)&  $.25$ ($p<.001$)\\ \hline  
         m. 4&  $.23$ ($p<.001$)&  $.26$ ($p<.001$)&  $.28$ ($p<.001$)&  $.28$ ($p<.001$)&  $-.05$ ($p=.269$)&  $.03$ ($p=.501$)\\ \hline  
         m. 5&  $-.05$ ($p=219$)&  $-.08$ ($p=.092$)&  $-.11$ ($p=.010$)&  $-.14$ ($p=.002$)&  $.09$ ($p=.042$)&  $.08$ ($p=.065$)\\ \hline  
 m. 6& $-.08$ ($p=.107$)&  $-.07$ ($p=.173$)&  $-.13$ ($p=.010$)&  $-.13$ ($p=.013$)&  $.08$ ($p=.126$)&  $.12$ ($p=.014$)\\ \hline  
 m. 7& $-.31$ ($p<.001$)&  $-.36$ ($p<.001$)&  $-.28$ ($p<.001$)&  $.28$ ($p<.001$)&  $-.01$ ($p=.846$)&  $-.01$ ($p=.790$)\\ \hline  
 m. 8& $-.13$ ($p=.008$)&  $-.17$ ($p<.001$)&  $-.22$ ($p<.001$)&  $-.20$ ($p<.001$)&  $.14$ ($p=.003$)&  $.11$ ($p=.028$)\\ \hline  
 m. 9& $.10$ ($p=.084$)&  $.09$ ($p=.127$)&  $.04$ ($p=.429$)&  $.06$ ($p=.310$)&  $.09$ ($p=.099$)&  $.14$ ($p=.015$) \\ \hline \end{tabular}
    \caption{Correlations between module presences and disease categories.}
    \label{tab:disease_correlations}
\end{table}

\subsubsection{Gene pathway analysis}
\begin{figure}[h]
     \centering
     \begin{subfigure}[b]{0.26\textwidth}
         \centering
         \includegraphics[width=\textwidth]{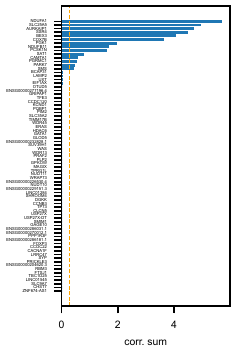}
         \caption{Module 1.}
     \end{subfigure}
     \hfill
     \begin{subfigure}[b]{0.26\textwidth}
         \centering
         \includegraphics[width=\textwidth]{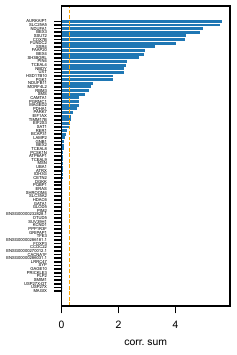}
         \caption{Module 2.}
     \end{subfigure}
     \hfill
     \begin{subfigure}[b]{0.26\textwidth}
         \centering
         \includegraphics[width=\textwidth]{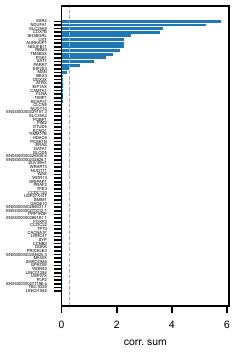}
         \caption{Module 3.}
     \end{subfigure}
     \hfill
     \begin{subfigure}[b]{0.26\textwidth}
         \centering
         \includegraphics[width=\textwidth]{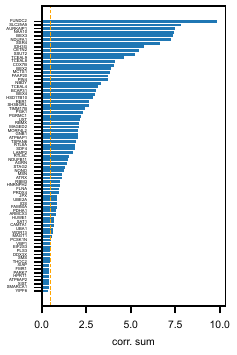}
         \caption{Module 4.}
     \end{subfigure}
     \hfill
     \begin{subfigure}[b]{0.26\textwidth}
         \centering
         \includegraphics[width=\textwidth]{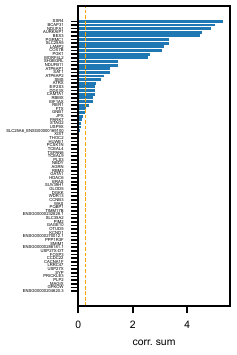}
         \caption{Module 5.}
     \end{subfigure}
     \hfill
     \begin{subfigure}[b]{0.26\textwidth}
         \centering
         \includegraphics[width=\textwidth]{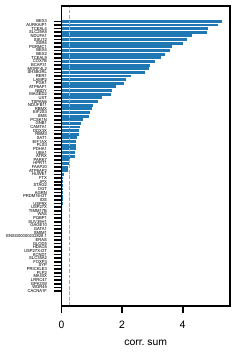}
         \caption{Module 6.}
     \end{subfigure}
     \hfill
     \begin{subfigure}[b]{0.26\textwidth}
         \centering
         \includegraphics[width=\textwidth]{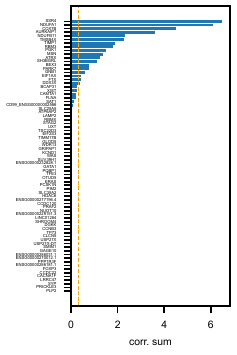}
         \caption{Module 7.}
     \end{subfigure}
     \hfill
     \begin{subfigure}[b]{0.26\textwidth}
         \centering
         \includegraphics[width=\textwidth]{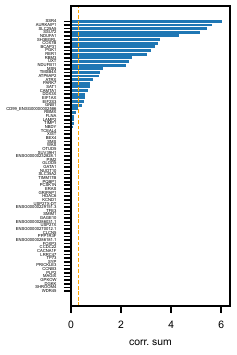}
         \caption{Module 8.}
     \end{subfigure}
     \hfill
     \begin{subfigure}[b]{0.26\textwidth}
         \centering
         \includegraphics[width=\textwidth]{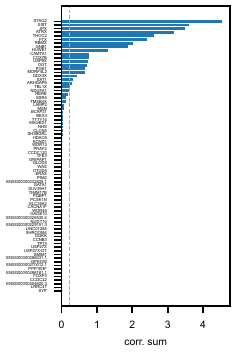}
         \caption{Module 9.}
     \end{subfigure}
     \caption{Top $75$ genes based on sums of correlations for each extracted module from the KPMP experiments. The orange line represents the cut-off line for gene subsets for ORA, equal to $5\%$ of the largest correlation sum.}
     \label{fig:kpmp_top_genes}
\end{figure}

\begin{figure}[h]
     \centering
     \begin{subfigure}[b]{0.33\textwidth}
         \centering
         \includegraphics[width=\textwidth]{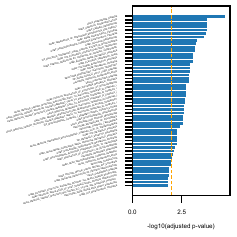}
         \caption{Module 1.}
     \end{subfigure}
     \hfill
     \begin{subfigure}[b]{0.33\textwidth}
         \centering
         \includegraphics[width=\textwidth]{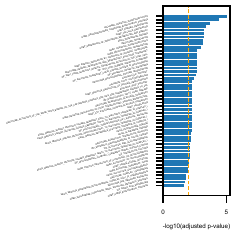}
         \caption{Module 2.}
     \end{subfigure}
     \hfill
     \begin{subfigure}[b]{0.33\textwidth}
         \centering
         \includegraphics[width=\textwidth]{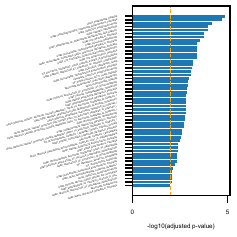}
         \caption{Module 3.}
     \end{subfigure}
     \hfill
     \begin{subfigure}[b]{0.33\textwidth}
         \centering
         \includegraphics[width=\textwidth]{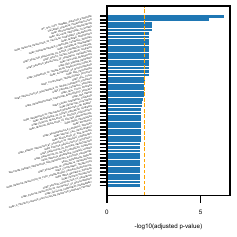}
         \caption{Module 4.}
     \end{subfigure}
     \hfill
     \begin{subfigure}[b]{0.33\textwidth}
         \centering
         \includegraphics[width=\textwidth]{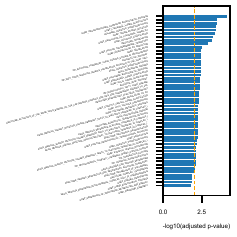}
         \caption{Module 5.}
     \end{subfigure}
     \hfill
     \begin{subfigure}[b]{0.33\textwidth}
         \centering
         \includegraphics[width=\textwidth]{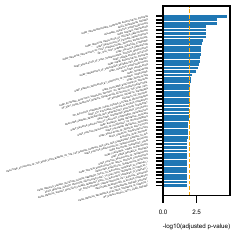}
         \caption{Module 6.}
     \end{subfigure}
     \hfill
     \begin{subfigure}[b]{0.33\textwidth}
         \centering
         \includegraphics[width=\textwidth]{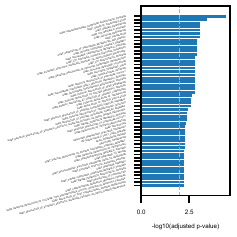}
         \caption{Module 7.}
     \end{subfigure}
     \hfill
     \begin{subfigure}[b]{0.33\textwidth}
         \centering
         \includegraphics[width=\textwidth]{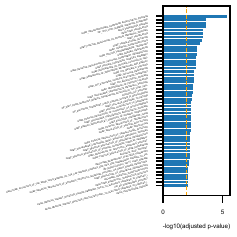}
         \caption{Module 8.}
     \end{subfigure}
     \hfill
     \begin{subfigure}[b]{0.33\textwidth}
         \centering
         \includegraphics[width=\textwidth]{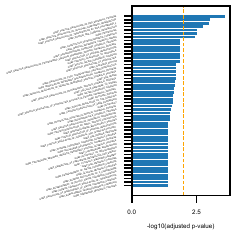}
         \caption{Module 9.}
     \end{subfigure}
     \caption{Top $50$ most significant pathways based on ORA for each of the $9$ extracted modules from the KPMP experiment. The orange line represents the significance $p$-value threshold of $0.01$, i.e., $2$ when looking at $-log_{10}(p)$ values.}
     \label{fig:kpmp_top_pathways}
\end{figure}

In Section \ref{sec:kpmp_exp}, we investigate the biological relevance of the patterns uncovered by LAVA based on the UMAP of the KPMP dataset. To do this, we perform overrepresentation analysis (ORA) of biological functions: a statistical method used to determine which functions are statistically overrepresented amongst the functional annotations of the genes in a set of interest (e.g. a LAVA module), compared to what would be expected by chance %ORA is as statistical method used to determine which pre-defined gene sets are more present in a subset of genes of interes than we would expect by chance  
\cite{huang_bioinformatics_2009}. %As pre-defined gene sets, 
We perform this analysis using three collections of functional annotations we thought were relevant for our disease context: (1) hallmark; (2) canonical pathways; and (3) the gene ontology biological process. These collections were downloaded from MSigDB \footnote{The Molecular Signatures Database, available at \hyperref[https://www.gsea-msigdb.org]{https://www.gsea-msigdb.org}.}. \textbf{Selection of genes per module.} To select the set of genes of interest for ORA, we first sum the correlations of each gene in a module, and then take the set of genes whose sum of correlations is %greater or equal to 
$\geq5\%$ of the largest sum %. We show that for all the modules in 
(Figure \ref{fig:kpmp_top_genes}). Accordingly, we performed ORA for each module using as background all measured genes from the KPMP dataset (a total of $31332$ genes). The resulting biological function $p$-values were adjusted using the Benjamini-Hochberg correction for multiple testing. \textbf{Results of ORA.} %Based on selected subsets of genes, 
We report the top 50 most significant results of ORA for each of the $9$ extracted modules in Figure \ref{fig:kpmp_top_pathways}. In the main paper, we highlight module $4$ because of its correlation with diseased samples, and CKD specifically (Section \ref{sec:kpmp_exp}, Table \ref{tab:disease_correlations}), which we think is of biological interest.

\end{document}